\documentclass{article}
\usepackage{amsmath,amsfonts,bm}

\def\eqreff#1{(\ref{#1})}
\def\eqref#1{equation~(\ref{#1})}
\def\1{\bm{1}}

\DeclareMathAlphabet{\mathsfit}{\encodingdefault}{\sfdefault}{m}{sl}
\SetMathAlphabet{\mathsfit}{bold}{\encodingdefault}{\sfdefault}{bx}{n}

\def\gA{{\mathcal{A}}}

\def\gG{{\mathcal{G}}}

\def\gL{{\mathcal{L}}}

\def\gS{{\mathcal{S}}}

\def\gX{{\mathcal{X}}}
\def\gY{{\mathcal{Y}}}

\def\sR{{\mathbb{R}}}

\newcommand{\E}{\mathbb{E}}

\newcommand{\mc}[3]{\multicolumn{#1}{#2}{#3}}
\newcommand{\mr}[2]{\multirow{#1}{*}{#2}}

\PassOptionsToPackage{numbers, compress}{natbib}
     \usepackage[preprint]{neurips_2023}

\usepackage[utf8]{inputenc} % allow utf-8 input
\usepackage[T1]{fontenc}    % use 8-bit T1 fonts
\usepackage{hyperref}       % hyperlinks
\usepackage{url}            % simple URL typesetting
\usepackage{booktabs}       % professional-quality tables
\usepackage{amsfonts}       % blackboard math symbols
\usepackage{nicefrac}       % compact symbols for 1/2, etc.
\usepackage{microtype}      % microtypography
\usepackage{xcolor}         % colors

\usepackage{amsmath}
\usepackage{amssymb}
\usepackage{mathtools}
\usepackage{amsthm}
\usepackage{bbm}
\usepackage{enumitem}
\usepackage{tabularx}
\usepackage{multirow}
\usepackage{diagbox}
\usepackage{wrapfig}

\usepackage[font=small,labelfont=bf]{caption}

\usepackage{algorithmicx}
\usepackage{algorithm}% http://ctan.org/pkg/algorithms
\usepackage{algpseudocode}% http://ctan.org/pkg/algorithmicx

\theoremstyle{plain}
\newtheorem{theorem}{Theorem}[section]

\newtheorem{lemma}[theorem]{Lemma}

\theoremstyle{definition}

\theoremstyle{remark}

\definecolor{darkblue}{rgb}{0.0,0.0,0.66}  
\definecolor{darkred}{rgb}{100,0.0,0.0} 
\hypersetup{colorlinks=true,linkcolor=darkred,citecolor=darkblue}

\newcommand{\vjnote}[1]{\textcolor{orange}{[VJ: #1]}}

\title{Debiasing Vision-Language Models \\ via Biased Prompts}

\author{%
  Ching-Yao Chuang$^\dagger$, Varun Jampani$^\ddag$, Yuanzhen Li$^\ddag$, \\ \textbf{Antonio Torralba$^\dagger$} \textbf{Stefanie Jegelka$^\dagger$}\\
  {$^\dagger$MIT CSAIL, $^\ddag$Google Research}
  \\
  \texttt{\{cychuang, torralba, stefje\}@mit.edu}
  \\
  \texttt{\{varunjampani, yzli\}@google.com} 
}

\begin{document}

\maketitle

\begin{abstract}
  Machine learning models have been shown to inherit biases from their training datasets. This can be particularly problematic for vision-language foundation models trained on uncurated datasets scraped from the internet. The biases can be amplified and propagated to downstream applications like zero-shot classifiers and text-to-image generative models. In this study, we propose a general approach for debiasing vision-language foundation models by projecting out biased directions in the text embedding. In particular, we show that debiasing only the text embedding with a calibrated projection matrix suffices to yield robust classifiers and fair generative models. The proposed closed-form solution enables easy integration into large-scale pipelines, and empirical results demonstrate that our approach effectively reduces social bias and spurious correlation in both discriminative and generative vision-language models without the need for additional data or training. The code is available at \url{https://github.com/chingyaoc/debias_vl}.
\end{abstract}

\section{Introduction}

Foundation vision-language models, such as CLIP \citep{radford2021learning}, DALLE-2 \citep{ramesh2022hierarchical}, Imagen \citep{saharia2022photorealistic}, and Stable Diffusion \citep{rombach2022high}, which are trained on extensive multimodal data at a massive scale, have led to a significant shift in the landscape of machine learning systems. Specifically, contrastive vision-language encoders like CLIP have the ability to perform zero-shot inferences without fine-tuning, and language embeddings can be used to train high-quality text-to-image models \citep{rombach2022high}.

While vision-language models demonstrate impressive capabilities, it is important to recognize that they may also exacerbate biases \citep{ mehrabi2021survey,agarwal2021evaluating, wang2021gender}. Recent studies \cite{birhane2021multimodal} have shown that the datasets these models are trained on can contain inappropriate image-text pairs with stereotypes, racist content, and ethnic slurs. The biases are then propagated to downstream applications \citep{agarwal2021evaluating, wang2021gender}, resulting in biased predictions. In addition to social biases, zero-shot models derived from vision-language models can also suffer from more general forms of spurious correlations such as image background, leading to poor group robustness \citep{zhang2022contrastive}. Biases also exist in generative models, where generated images may exhibit bias towards certain genders and races \citep{cho2022dall, mishkin2022dall}. Substantial progress has been made recently toward mitigating biases in vision-language models \citep{parraga2022debiasing, berg2022prompt, zhang2022contrastive}. However, many current approaches for addressing bias in models require training or fine-tuning the models using resampled datasets or modified objectives, which can be computationally intensive for foundation models.

In this work, we propose a general approach for self-debiasing foundation vision-language models by projecting out biased directions in the text embedding. Given a vision-language encoder such as CLIP, we define a set of biased directions in the embedding using prompts that describe the biases. For instance, prompts like ``a photo of a male/female'' define a biased subspace in the latent space. One approach to mitigating these biases is to construct a projection matrix, a linear transformation of the text embedding that projects out the biased directions \citep{bolukbasi2016man}. However, solely relying on prompts to define biased directions may be unstable and noisy \citep{gonen2019lipstick}. To address this issue, we propose a calibration loss that minimizes the discrepancy of a pair of prompt embeddings. For example, given a projection matrix that removes gender information, the projected vectors of prompts ``a photo of a male doctor'' and ``a photo of a female doctor'' should be similar. Based on this principle, we design an objective to calibrate the projection matrix, which has an easily solvable closed-form solution. This allows for the construction of the projection matrix to be \emph{training-free and requires no downstream dataset or labels}, making it suitable for large-scale models. Empirically, we find that debiasing only the text embedding with a calibrated
projection matrix suffices to improve the group robustness of zero-shot models on well-established benchmarks.

We then extend our approach to generative models such as Stable Diffusion \citep{rombach2022high}, a widely adopted text-to-image model conditioned on text embeddings from CLIP \citep{radford2021learning}. The inherent challenge lies in the fact that generative models are distinctly dissimilar from zero-shot classification, where the target classes are explicitly defined. With generative models, our objective is to develop a debiasing matrix that is universally applicable to every prompt. This matrix can then be employed as a standard preprocessing stage prior to feeding the text embedding into the generative model. To accomplish this, we solve the calibration matrix with a set of positive pairs which comprise various prompts from the training dataset, and debias the unseen prompts with the obtained matrix. Similar to debiasing zero-shot models, the projection matrix improves the diversity of generated images from text-to-image models without altering the model parameters.

In short, this work makes the following contributions:
\vspace{-3mm}
\begin{itemize}[leftmargin=0.5cm]\setlength{\itemsep}{-1pt}
\item We present a simple and general approach for debiasing vision-language models; %by robustly projecting out biases in the text embedding;
\item The proposed approach does not require training, data, or labels, making it computationally efficient for use with foundation models;
\item We evaluate our approach through experiments on both discriminative (zero-shot, text-image retrieval) and generative (diffusion) vision-language models.
\end{itemize}

\section{Related Works}

Vision-Language models \citep{radford2021learning, ramesh2022hierarchical, saharia2022photorealistic, rombach2022high} have become increasingly widespread in recent years. However, these models are known to suffer from spurious correlations and can be biased towards certain races and gender. \citet{birhane2021multimodal} study the datasets these models are trained on and show that their biases can be inherited by the models. Various methods have been proposed to address biases, but many of them only address single-modality models.

\paragraph{Biases in Language Models}
Large-scale language models have been shown to contain harmful or misrepresentative biases  \citep{blodgett2020language, nadeem2020stereoset,weidinger2021ethical}. Previous research has demonstrated the presence of gender bias in natural language processing systems \citep{bolukbasi2016man, zhao2019gender} as well as racial bias \citep{manzini2019black, garg2018word}. \citet{bolukbasi2016man} first proposed the use of orthogonal projections to remove gender biases in word embeddings. This approach was later extended to debiasing sentence embeddings \citep{liang2020towards}. %However, it has been shown that solely relying on orthogonal projection may not fully remove gender biases \citep{gonen2019lipstick}. %\sj{but you re also just doing that? CY: We introduce the calibration loss to improve it.} 
Alternative methods include regularizing the models with constraints on training data \citep{zhao2017men, huang2019reducing} or directly modifying the dataset \citep{sun2019mitigating, zhao2019gender}. However, scaling these approaches to large foundation models can be challenging as they often require retraining the backbone encoders.

\paragraph{Biases in Vision Models}
Gender and racial biases have also been widely explored in computer vision \citep{alvi2018turning, wang2020mitigating}, in terms of dicsriminative models \citep{wang2019racial} and generative models \citep{xu2018fairgan, grover2019bias, cho2022dall}. Many debiasing approaches aim to learn good representations via adversarial training \citep{madras2018learning, wang2020towards}, or augmenting the biased dataset \citep{ramaswamy2021fair, chuang2021fair}. Beyond social bias, many works study spurious correlations, a more general form of bias that can include features such as image background or other non-target attributes that are correlated with labels. This problem of spurious correlations is often studied and tackled as a group robustness problem \citep{sagawa2019distributionally, izmailov2022feature}. \citet{kirichenko2022last} show that last layer re-training is sufficient for robustness to spurious correlations, which aligns with our finding that debiasing the zero-shot weights suffices to yield robust classifiers. %\vjnote{The last sentence is not clear at the point and requires more context to understand.} \cy{fix it with more explanations.}

\paragraph{Biases in Vision-Language Models}
Recently, biases in multimodal settings have gained significant attention \citep{agarwal2021evaluating, hall2023vision}. \citet{wang2021gender} propose to remove dimensions %\sj{really coordinates? that's simplistic... CY: Yes...} 
in the CLIP embedding that are highly correlated with gender attributes. \citet{berg2022prompt} debias the CLIP models with prompt learning via an adversarial approach. \citet{seth2023dear}  learn
additive residual image representations to offset the biased representations. Recently, \citet{zhang2022contrastive} address the group robustness of vision-language models with contrastive learning. These previous works are data-oriented, where models are trained or finetuned on labeled datasets. In contrast, our approach is fully zero-shot, which does not require any downstream dataset and model training. To debias generative models, a recent work \citep{friedrich2023fair} pre-defines a look-up table to provide fair guidance for text-to-image diffusion models. Nevertheless, this method encounters limitations when faced with previously unseen classes that are absent from the look-up table, while our approach generalizes well to new concepts.

%Beyond vision-language setting, biases of machine learning models has also been studied extensively in fairness community \citep{barocas-hardt-narayanan}, where rigorous definitions and provable algorithms are proposed. 

\section{Biases and Spurious Correlations}

%Say that there is a line of work that focuses on debiasing the last layer, and we follow that line of approach.
We consider a dataset in which each input $x \in \gX$ is associated with multiple attributes, including the target class $y \in \gY$ and a spurious attribute $a \in \gA$. We focus on the case where biases are present and the attribute $a$ is spuriously correlated with the label $y$. For instance, the  class ``doctor'' could be correlated with the spurious attribute ``gender'' in the datasets foundation models are trained on \citep{birhane2021multimodal}. Importantly, these biases can be transferred to downstream tasks, both discriminative and generative.

\paragraph{Discriminative Models}
In this work, we examine the biases present in zero-shot classifiers obtained via a vision-language encoder such as CLIP. These classifiers are built by assigning each row of the linear classifier weight $\beta \in \sR^{K \times d}$ to be the embedding of a ``class prompt'', for example, ``a photo of a [class name]'' \citep{radford2021learning}. Importantly, it does not require any data or training to construct these zero-shot classifiers. However, it is possible for these zero-shot classifiers to inherit biases from the dataset used to train the vision-language models. To study these biases, we utilize the group robustness framework proposed by \citet{sagawa2019distributionally}. In this setting, groups are defined by a combination of the labels and spurious attributes: $\gG \in \gY \times \gA$. Given a distribution $P_g$ conditioned on $g \in \gG$ and a loss function $\ell: \gY \times \gY \rightarrow \sR$, group robustness requires that the classifier $f: \gX \rightarrow \gY$ achieves a small gap between its worst-group error and average error:
\begin{align}
    \label{eq_obj_discriminative}
   \max_{g \in \gG}\E_{x, y \sim P_g} \left[ \ell(f(x), y) \right] - \E_{x, y \sim P} \left[ \ell(f(x), y) \right].
\end{align}

The definition of metrics for text-image retrievals, such as maximal skewness \citep{geyik2019fairness}, will be deferred to the experiment section.

\paragraph{Generative Models}
A text-to-image model learns a conditional distribution $\hat{P}(X | Z = z)$, where $z$ is the embedding of the prompt. However, the biased nature of the dataset used to train the generative model can affect the distribution $\hat{P}$. To measure the bias present in generative models, recent works \citep{choi2020fair, teo2021measuring} propose using statistical parity. Specifically, given a classifier $h: \gX \rightarrow \gA$ for the spurious attribute, the discrepancy of the generative distribution $\hat{P}$ is defined as the L2 norm between empirical and uniform distributions \citep{choi2020fair}:
\begin{align}
\label{eq_obj_generative}
    \sqrt{\sum_{a \in \gA}  \left( \E_{x \sim \hat{P}} \left[ \mathbbm{1}_{h(x) = a} \right] - 1/|\gA| \right)^2}
\end{align}
In practice, the expectation is estimated with empirical samples. A fair generative model minimizes the discrepancy by ensuring that each attribute $a \in \gA$ has an equal probability (uniformly distributed).

\section{Debiasing Discriminative Models}

It is essential for a robust classifier to evade dependence on irrelevant features present in images. This necessitates the classifier to be invariant to image backgrounds and insensitive to attributes such as race or gender. Prior research has employed datasets with target labels and spurious attributes %\sj{do you mean there are labels for attributes? Make that clear in the way you are saying it} 
to quantify and eliminate biases \citep{sagawa2019distributionally, zhang2022contrastive}. However, this approach is not feasible in a zero-shot setting, where data and training are prohibitive. %\sj{prohibitively large?} \cy{zero-shot models does not need data or training. I add a line in the previous section to clarify this.}

\subsection{Measuring Biases with Prompts}
\label{sec_spurious_prompt}

\begin{figure}[t]
\begin{minipage}{\textwidth}
\vspace{-1mm}
  \begin{minipage}[b]{0.42\textwidth}
    \centering
    \begin{tabularx}{\textwidth}{l | c c |  c c }
\toprule
%\multirow{2}{*}{\diagbox{y}{a}}
&  \multicolumn{2}{c}{\footnotesize \textbf{CelebA}} & \multicolumn{2}{c}{\footnotesize \textbf{Waterbird}}  \\
 %\cmidrule(lr){2-3} \cmidrule(lr){4-5} 
  & \footnotesize Male & \footnotesize Female & \footnotesize Land  & \footnotesize Water  \\
 \midrule
 $\beta_{y=0}$  &  0.83 &  0.78 &  0.75  & 0.66
\\
 $\beta_{y=1}$  &  0.77 &  0.85 &  0.65 &  0.70
\\
\bottomrule
\end{tabularx}
      \captionof{table}{\textbf{Cosine similarity between classifier weights and spurious directions.} %\vjnote{Text in 4.1 says the cosine similarity with text embedding (not classifier weights)? cy: they are the same thing, where I feel like the classifier weight might give more info here} 
      In both datasets, the classifier weights are biased toward certain spurious attributes. %\vjnote{The table presentation is a bit strange with attributes from different datasets combined together. Better to separate them. Also, tables usually look better without vertical separations. Is it better to use cmidrule to group columns like I edited for this table for other tables? Use capitals for the first letters in the words?}}
      %The labels and spurious attributes are binary variables in both datasets.
      }
      \label{table_cos}
    \end{minipage}
    \hfill
  \begin{minipage}[b]{0.55\textwidth}
    \centering
    \includegraphics[width=\linewidth]{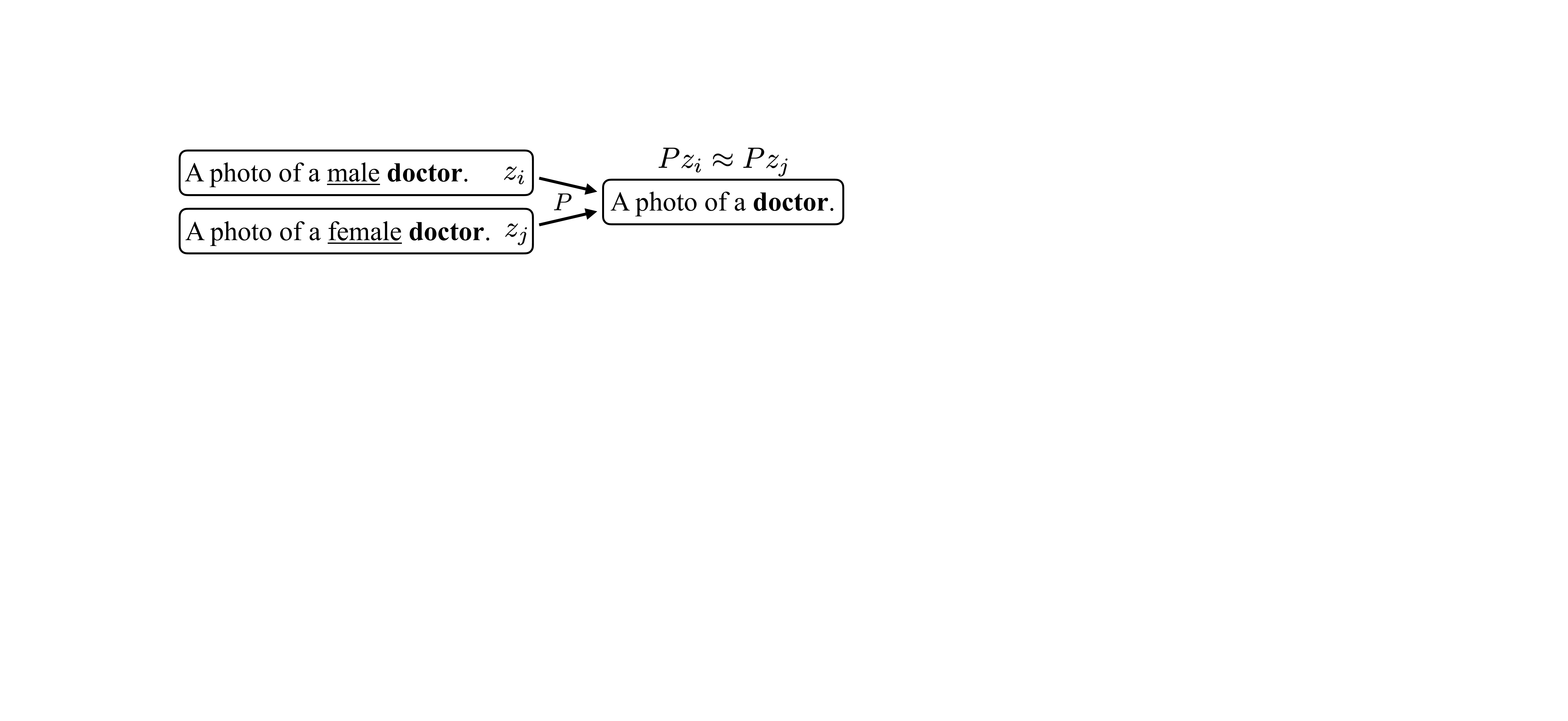}
    \captionof{figure}{\textbf{Calibration with Positive Pairs.} Upon projecting out irrelevant features (such as gender), the embeddings of group prompts should exhibit similarity and contain only information pertaining to the target class (e.g. doctor).}
    \label{fig_calibrate}
  \end{minipage}
\end{minipage}
\end{figure}

In contrast to previous approaches, our proposed method for measuring biases utilizes prompts, drawing inspiration from studies on debiasing word embeddings \citep{bolukbasi2016man}. The use of vision-language contrastive training allows for the description of irrelevant features through natural language. As such, embeddings of prompts such as ``a photo of a [irrelevant attribute]" can capture these spurious features in the visual embedding. Consequently, the bias of a classifier can be quantified by computing the cosine similarity between its weights and the corresponding spurious feature. %\sj{there is no spurious embedding, just spurious features. Be explicit how you get the direction that you use in the inner product} 
Table \ref{table_cos} illustrates the cosine similarity between the embeddings of prompts that describe the target classes and irrelevant attributes, %\sj{same, what is a spurious prompt? prompt for spurious attribute? (at most...)} 
using two popular group robustness benchmarks: Waterbird \citep{sagawa2019distributionally} and CelebA \citep{liu2015deep}. The details of datasets and the specific prompts can be found in section \ref{sec_exp} and appendix \ref{sec_prompts}. The results demonstrate that the classifier weights are inclined towards certain irrelevant attributes (gender or image background), implicitly implying that the classifiers are using these spurious directions to make predictions.

\subsection{Debiasing via Orthogonal Projection}
\vspace{-2mm}
As the zero-shot weights can also be viewed as natural language embeddings, a straightforward approach is to follow the debiasing pipeline employed in word and sentence embeddings \citep{bolukbasi2016man, liang2020towards}. In particular, to make the classifier invariant to these irrelevant features, we align the classifier weights with the orthogonal complement of these embeddings. Let $A \in \sR^{d \times m}$ be a matrix whose columns are the embeddings of spurious prompts. %\vjnote{Better to make it explicit what we mean by `spurious directions as columns'. They are text embeddings of spurious attributes? Explaining with an example would be even better as this helps in better understanding the motivation in 4.3.} 
The orthogonal projection matrix is then:
\begin{align*}
    P_0 = I - A(A^T A)^{-1} A^T.
\end{align*}
We can use the projection matrix to eliminate spurious directions in a text embedding $z$ as $P_0 z$.

%\sj{use either $P\beta$ or $\beta P$ uniformly throughout. See Figure 2. Actually, using $\beta$ for the embedding is a bit confusing, since $\beta$ is by standard the classifier/regression weights and not the data. Why not $z$ or sth like that?}\sj{I see now where the confusion comes from, since $\beta$ is both data and classifier.}

\subsection{Calibrating the Projection Matrix}
\vspace{-2mm}
\label{sec_cali}
It is essential to acknowledge that the estimation of the irrelevant feature directions may introduce an approximation error in the projection matrix \citep{gonen2019lipstick}. Additionally, in certain scenarios, it may be challenging to thoroughly describe the irrelevant attribute using a limited number of prompts, resulting in increased uncertainty in the projection matrix estimation. This issue is also evident in our empirical results (Table \ref{table_group_robust} and \ref{table_ablation}), where the use of orthogonal projection fails to enhance performance.
%\sj{point to where these results are shown, if they are included? May be good to include them as an ``ablation'' for the calibration.}

%\paragraph{[TODO] Emphasize that class-conditioned prompts are essential.}

To improve the estimation of the projection matrix, we leverage \emph{positive pairs} of prompts that are expected to have the same semantic meaning after projection. In particular, the embedding of prompts such as ``a photo of a [class name] with [spurious attribute]'' should only contain information about ``[class name]'' after projecting out the spurious information, as Figure \ref{fig_calibrate} illustrates. Motivated by this intuition, we propose to regularize the difference between the projected embeddings using a set of positive pairs $S$:
\begin{align}
 \min_{P} \;  \left\| P - P_0 \right\|^2 + \frac{\lambda}{|\gS|} \sum_{(i,j) \in \gS} \left \| P z_i -  P z_j \right\|^2,
 \label{eq_cali}
\end{align}
where $(z_i, z_j)$ is the embedding of pair $(i,j)$ in $\gS$ and $(i, j)$ are prompts that describe the same class but different spurious attributes. The loss encourages the linear projection $P$ to be invariant to the difference between $(i,j)$, i.e., the spurious attributes. The optimization problem has a convenient closed-form solution, as demonstrated in Lemma \ref{lemma_calibrate}. %\sj{say that $i,j$ are from the same class. This will though also forget all other things that make。$i,j$ different and may be unrelated to the spurious attribute.}
\begin{lemma}
The minimizer of the calibration loss is
\begin{align*}
P^\ast = P_0 \Bigl(I + \frac{\lambda}{|\gS|} \sum_{(i,j) \in \gS} (z_i - z_j)(z_i - z_j)^T\Bigl)^{-1}.
\end{align*}
\label{lemma_calibrate}
\vspace{-4mm}
\end{lemma}

We can obtain an interpretation of the minimizer by relating it to singular value decomposition (SVD). Let $Z_\textnormal{diff} \in \mathbb{R}^{d \times |S|}$, where the columns of $Z_\textnormal{diff}$ enumerate the pairwise difference $z^i - z^j$ for all $(i, j) \in S$. The matrix $Z_\textnormal{diff}$ defines a subspace that represents the variation in the embedding when the irrelevant feature is changed. Using $Z_\textnormal{diff}$, the minimizer can be written as $P^\ast = P_0 (I + \lambda' Z_\textnormal{diff} Z_\textnormal{diff}^T)^{-1}$, where we define $\lambda' = \lambda/|\gS|$ to simplify the notation. Assume that the SVD of $Z_\textnormal{diff}$ is $U \Sigma V^T$. Then we have $Z_\textnormal{diff} Z_\textnormal{diff}^T = U \Sigma^2 U^T$. The optimal solution $P^\ast$ can then be rewritten as
\begin{align*}
    P^\ast &= P_0(U(I + \lambda' \Sigma^2)U^{T})^{-1} =P_0 \underbrace{U(I + \lambda' \Sigma^2)^{-1}U^{T}}_{\textnormal{Calibration Matrix}}.
\end{align*}
We can see that $U(I + \lambda' \Sigma^2)^{-1}U^{T}$ acts as a calibration term. Before multiplying the text embedding with the projection matrix $P_0$, %\sj{wait - are you multiplying the classifier or the data points? be consistent in that!}
variation due to the change of the spurious feature, namely, the eigenvectors with large squared singular value in $Z_\textnormal{diff}$ (spurious direction) will be down-weighted due to the inverse $(I + \lambda' \Sigma^2)^{-1}$. Therefore, varying the spurious attributes should result in similar embeddings after multiplying the calibration matrix. %\sj{I don't yet fully get that, why does rescaling the eigenvalues help when you vary the directions?} \cy{modify a few words to make this more clear}

%\vspace{-1mm}
\subsection{Relation to an Equalization Loss}
%\vspace{-1mm}

Finally, we provide an equivalent form of the calibrated projection and relate it to an equalization loss. Ideally, we want each row of the classifier weight $\beta \in \sR^{K \times d}$ to have similar cosine similarity to pairs of embeddings in $\gS$. For instance, the embedding of ``a photo of a doctor'' should be equally similar to ``a photo of a male doctor'' and ``a photo of a female doctor''. In this section, we will show that the optimum of the calibration loss does satisfy this criterion.

%Let $\beta \in \sR^{K \times d}$ be the classifier weight of zero-shot models and $(z_i, z_j)$ be the embedding of positive pair $(i,j)$ in $\gS$, we consider the following objective 

%\sj{again, don't use the same notation for classifier weight and data, it's confusing. I guess you are using the embedding as classifier - as I infer implicitly, but I still couldn't follow that easily.}

We consider the following objective for obtaining a debiased text embedding $z \in \sR^d$ of a prompt given its initialization $z_0 \in \sR^d$ from the text encoder:
\begin{align}
\min_{z} \;  \| z - z_0 \|^2 + \frac{\lambda}{|\gS|} \sum_{(i,j) \in \gS} ( z^T z_i -  z^T z_j )^2.
\label{eq_equal}
\end{align}
The loss encourages the embedding $z$ to have similar cosine similarity to embeddings in positive pairs while maintaining proximity to the initialization $z_0$. Objective \eqreff{eq_equal} has the same optimal solution as the calibration loss \eqreff{eq_cali}. 
\begin{lemma}
The minimizer of objective \eqreff{eq_equal} reads
\begin{align*}
z^\ast = \underbrace{\Bigl( I + \frac{\lambda}{|\gS|} \sum_{(i,j) \in \gS} (z_i - z_j)(z_i - z_j)^T \Bigl)^{-1}}_{\textnormal{Calibration Matrix}} z_0
\end{align*}
In particular, we have $P_0 z^\ast = {P^\ast} z_0$ where $P^\ast$ is the minimizer of the calibration loss \eqreff{eq_cali}.
\label{lemma_equivalent}
\end{lemma}

Lemma \ref{lemma_equivalent} shows that the optimal solution of \eqreff{eq_equal} is equivalent to multiplying the original embedding $z$ with the calibration matrix defined before. Applying the projection $P_0$ to $z^\ast$ leads to the same weight in Lemma \ref{lemma_calibrate}. This interpretation is particularly useful in cases where the ideal solution does not lie in the middle of $z_i$ and $z_j$, as will be shown in section~\ref{sec:generative} where we address biases in generative models.

The equalization objective has a similar motivation as the equalization step proposed by \citet{bolukbasi2016man} in their work on removing gender bias from word embeddings. Similar to the idea of positive pairs, given a set of word embeddings that has the same semantic meaning except for gender, their approach centers these embeddings by setting them to the average embedding of the set. After centering, any word in the dictionary will be equidistant to all words in the set. However, our approach differs in that we modify the embedding of the target prompt $z$, rather than the embedding of positive pairs, making it more suitable for debiasing zero-shot classifiers as we are primarily concerned with the embedding of $z$.

%\sj{biased?} In particular, their approach modifies the embedding of positive pairs $(z_i,  z_j)$ instead of $z$ itself, which is not applicable for debiasing zero-shot classifiers. \sj{is the  bolukbasi et al approach then equivalent to eqn 3? that may be a novelty concern, but if it is the case, then be upfront about it. Just don't claim it as your contribution then.}
\vspace{-1mm}
\section{Experiments: Discriminative Models}
\label{sec_exp}
\vspace{-1mm}

\begin{table}[t]
\small
\begin{center}{%
\begin{tabularx}{\textwidth}{l | c c c | c c c | c c c | c c c }
\toprule
Backbone &  \multicolumn{6}{c}{CLIP ResNet-50} & \multicolumn{6}{c}{CLIP ViT-L/14} \\
\midrule
Dataset &  \multicolumn{3}{c}{\textbf{Waterbird}} & \multicolumn{3}{c}{\textbf{CelebA}} &  \multicolumn{3}{c}{\textbf{Waterbird}} & \multicolumn{3}{c}{\textbf{CelebA}}  \\
 & WG & Avg & Gap & WG & Avg & Gap & WG & Avg & Gap & WG & Avg & Gap  \\
 \midrule
\mc{5}{l}{\textit{methods using data and labels}} \\ 
ERM Linear & 7.9 & 93.5 & 85.6 & 11.9 & 94.7 & 82.8 & 65.9 & 97.6 & 31.7 & 28.3 & 94.7 & 66.4
\\
ERM Adapter & 60.8 & 96.0 & 35.2 & 36.1 & 94.2 & 58.1 & 78.4 & 97.8 & 19.4 & 36.7 & 94.2 & 57.5
\\
WiSE-FT & 49.8 & 91.0 & 41.2 & 85.6 & 88.6 & 3.0  & 65.9 & 97.6 & 31.7 & 80.0 & 87.4 & 7.4
\\
DFR (Sub) & 63.9 & 91.8 & 27.9 & 76.9 & 92.5 & 15.6 & 51.9 & 95.7 & 43.8 & 76.3 & 92.1 & 15.8 
\\
DFR (Up) & 51.3 & 92.4 & 41.1 & 89.6 & 91.8 & 2.2 & 65.9 & 96.1 & 30.2 & 83.7 & 91.2 & 7.5
\\
CA  &  83.7 & 89.4 & 5.7 & 90.0 & 90.7 & 0.7 &  86.9 & 96.2 & 9.3 & 84.6 & 90.4 & 5.8
\\
\midrule
\mc{5}{l}{\textit{methods without data and labels}} \\  
Zero-shot  & 39.6 & 77.3 & 37.7  & 75.9 & 82.3  & 6.4  & 45.3 & 84.4 & 39.1  & 72.8 & 87.6  & 14.9
\\
Orth-Proj (Ours) & 48.1 & 83.6 & 35.4 & 61.4 & 86.4 & 25.0 & 61.4 & 86.4 & 25.0 & 71.1 & 87.0 & 15.9
\\
Orth-Cali (Ours) & \textbf{74.0} & 78.7 & \textbf{4.7} & \textbf{82.2} & 84.4 & \textbf{2.2} & \textbf{68.8} & 84.5 & \textbf{15.7} & \textbf{76.1} & 86.2 & \textbf{10.1}
\\
\bottomrule
\end{tabularx}}
\end{center}
%\vspace{-1mm}
\caption{\textbf{Group Robustness of Vision-Language Models.} For each backbone, the first blocks contain methods that require data and labels, while the second blocks contain zero-shot methods. The numbers for the first block are adopted from \citet{zhang2022contrastive}. The proposed calibration loss achieves comparable or even smaller gaps between average and worst group accuracy without the need for any data or labels.
}
\vspace{-5mm}
\label{table_group_robust}
%\vspace{-3mm}
\end{table}

%We now evaluate our approach with experiments in discriminative (zero-shot classifier, text-image retrieval) and generative (text-to-image) models. 

%\vspace{-1mm}
\subsection{Group Robustness against Spurious Correlations}
%\vspace{-1mm}

By following the setting of \citet{zhang2022contrastive}, we evaluate our approach on two popular benchmarks for evaluating spurious correlations, Waterbird \citep{sagawa2019distributionally} and CelebA \citep{liu2015deep}. On Waterbird, a water/land background is a confounding factor for the waterbirds/landbirds class, while on CelebA the binary gender is the spurious feature for blond/dark hair. Therefore, both datasets contains four groups defined by the labels and the spurious attributes. As such, both datasets contain four groups defined by the labels and the spurious attributes. 

We evaluate our approach against several baselines, including zero-shot classification \citep{radford2021learning}, empirical risk minimization (ERM) with linear probing \citep{kumar2022fine}, and ERM with non-linear adapter \citep{gao2021clip}. Additionally, we also consider three recent methods designed to improve the group robustness of vision-language foundation classifiers:
%\vspace{-2mm}
\begin{itemize}[leftmargin=0.5cm]\setlength{\itemsep}{-1pt}
\item Weight Space Ensembling (WiSE-FT) \citep{wortsman2022robust}, which trains a linear classifier first using ERM and then combines the classifier outputs with the initial zero-shot predictions;
\item Deep Feature Reweighting (DFR) \citep{kirichenko2022last}, which trains a linear probe on embeddings obtained from a pre-trained model using group-balanced data. Following \citet{zhang2022contrastive}, the group labels are replaced with zero-shot predictions;
\item Contrastive Adapter (CA) \citep{zhang2022contrastive}, which trains adapters using contrastive learning to bring embeddings in the same class closer.
\end{itemize}

It is important to note that \textbf{all of the baselines}  except the zero-shot classifier \textbf{require at least training data and class labels}, while our debiasing approach does not require access to any input data, labels, or group labels, which follows the principles of zero-shot learning.

%------------------------------------------

\begin{wraptable}{r}{6.5cm}
\vspace{-4mm}
\setlength\tabcolsep{3.5pt}
\begin{tabular}{l | c c c | c c c }
\toprule
&  \multicolumn{3}{c}{\small \textbf{Waterbird}} & \multicolumn{3}{c}{\small \textbf{CelebA}}  \\
 & \small WG & \small Avg & \small Gap & \small WG & \small Avg & \small Gap  \\
 \midrule
$\lambda = 200$ & 71.8 & 80.8 & 9.0 & 80.7 & 83.9 & 3.2
\\
$\lambda = 400$& 72.9 & 79.5 & 6.6 & 81.6 & 84.2 & 2.6
\\
$\lambda = 600$& 73.5 & 79.2 & 5.7 & 81.9 & 84.3 & 2.4
\\
$\lambda = 1000$ &74.0 & 78.7 & 4.7 & 82.2 & 84.4 & 2.2
\\
\bottomrule
\end{tabular}
\caption{\textbf{Sensitivity to $\lambda$.} We vary the weighting parameter $\lambda$ and evaluate group robustness with ResNet-50 backbone.}\label{table_lambda}
\vspace{-2.5mm}
\end{wraptable} 

We evaluate the performance of our proposed approach using two CLIP backbones: ResNet-50 \citep{he2016deep} and ViT-L/14 \citep{dosovitskiy2020image}. The results are presented in Table \ref{table_group_robust}. The results indicate that a simple application of the orthogonal projection (Orth-Proj) by itself only yields limited improvement of the worst group accuracy, whereas the calibration loss (Orth-Cali) significantly improves robustness across datasets and base models. The proposed Orth-Cali method achieves comparable or even smaller gaps between average and worst group accuracy compared to the state-of-the-art contrastive adapter \citep{zhang2022contrastive}, without the need for any data or labels. Note that the baselines generally achieve better average accuracy as they require fine-tuning on the target datasets.

Empirically, we found that gradually increasing the parameter $\lambda$ improves the worst group accuracy and leads to a stable solution as shown in Table \ref{table_lambda}. Therefore, for all the experiments on discriminative models, we set $\lambda$ to $1000$ by default. To investigate the importance of orthogonal projection and calibration, we present an ablation study in Table \ref{table_ablation}. The results indicate that the calibration loss alone ($P_0 = I$) performs well on the CelebA dataset, as the spurious feature (gender) is relatively easy to describe with prompts. However, performance drops on the Waterbird dataset without a good initialization from the orthogonal projection. More ablation studies can also be found in Appendix \ref{appendix_exp_results}, where we demonstrate the importance of class names in positive pairs.

\subsection{Debiased Information Retrieval}
\vspace{-2mm}
Fairness in text-image retrieval has gained increasing attention in recent years. Building on the work of \citet{berg2022prompt}, we propose to utilize the MaxSkew metric, introduced by \citet{geyik2019fairness}, to evaluate the level of fairness in the retrieval results. Specifically, we conduct our analysis on the FairFace dataset \citep{karkkainen2019fairface}, which is specifically designed to address issues of fairness in facial recognition systems. Given a ranked list of images in response to a text query, let $r_{a,k}$ be the ratio of the top k images that are labeled with attribute $a$. Then MaxSkew@k is defined as $\max_{a \in \gA} \log \frac{r_{a,k}}{1 / |\gA|}$.
It quantifies the maximal discrepancy between the ratio of top k images labeled with a specific sensitive attribute, denoted as $r_{a,k}$, and the uniform weight $1 / |\gA|$, where $\gA$ represents the set of sensitive attributes. The MaxSkew metric provides a useful measure of fairness in text-image retrieval systems, by assessing the degree to which the retrieval results are evenly distributed across sensitive attributes. A small MaxSkew value indicates that the distribution of retrieved images across different sensitive attributes is close to being uniform.

\begin{figure}[!t]
\begin{minipage}{\textwidth}
\vspace{-1mm}
  \begin{minipage}[b]{0.49\textwidth}
    \centering
    \setlength\tabcolsep{3.5pt}
    \begin{tabularx}{\textwidth}{l | c c c | c c c }
\toprule
&  \multicolumn{3}{c}{\small \textbf{Waterbird}} & \multicolumn{3}{c}{\small \textbf{CelebA}}  \\
 & \small WG & \small Avg & \small Gap & \small WG & \small Avg & \small Gap  \\
 \midrule
Proj only & 48.1 & 83.6 & 35.4 & 61.4 & 86.4 & 25.0
\\
Cali only & 55.6 & 81.6 & 26.0 & 81.6 & 84.7 & 3.1
\\
Proj + Cali & \textbf{74.0} & 78.7 & \textbf{4.7} & \textbf{82.2} & 84.4 & \textbf{2.2}
\\
\bottomrule
\end{tabularx}
\vspace{-1mm}
      \captionof{table}{\textbf{Dissecting Orthogonal Projections.} We evaluate variants of orthogonal projection with ResNet-50 backbone.}% The labels and spurious attributes are binary variables in both datasets.}
      \label{table_ablation}
    \end{minipage}
    \hfill
  \begin{minipage}[b]{0.49\textwidth}
    \centering
    \setlength\tabcolsep{3.5pt}
    \begin{tabularx}{\textwidth}{l | c c c | c c c }
\toprule
&  \multicolumn{3}{c}{\small \textbf{CLIP ViT-B/32}} & \multicolumn{3}{c}{\small \textbf{CLIP ViT-L/14}}  \\
 & \small Gen & \small Race & \small Age & \small Gen & \small Race & \small Age  \\
 \midrule
Zero-shot  & .206 & .743 & .797  & .206 & .768  & .703
\\
Orth-Proj & .146 & .755 & \textbf{.635} & .349  & .605 & .706
\\
Orth-Cali & \textbf{.102} & \textbf{.638} & .641 & \textbf{.200}  & \textbf{.461} & \textbf{.662}
\\
\bottomrule
\end{tabularx}
\vspace{-1mm}
    \captionof{table}{\textbf{Measuring biases on FairFace.} We MaxSkew@1000 (the smaller the better) on FairFace validation set.}
    \label{table_fairface}
  \end{minipage}
\end{minipage}
\vspace{-4mm}
\end{figure}

To measure the bias, we query the validation set of FairFace based on 10 prompts that are uncorrelated with facial expressions or sensitive attributes, e.g., ``a photo of a [concept] person'', where the [concept] is a neutral concept such as evil or smart. The detailed prompts are described in Appendix \ref{appendix_exp_details}. We measure the MaxSkew based on three labeled attributes of FairFace: gender, race, and age. Table \ref{table_fairface} shows the average MaxSkew@1000 over concepts, demonstrating that our approach significantly reduces the MaxSkew across different attributes and backbones.

\vspace{-1mm}
\section{Debiasing Generative Models}\label{sec:generative}
\vspace{-1mm}

We now explore the possibility of extending the methodology developed for discriminative models to generative models. Our primary focus is on addressing social group biases, specifically gender and race discrepancy, as measured by metric \eqreff{eq_obj_discriminative}. In particular, the main experiment is to query the generative model using profession-related prompts, specifically ``a photo of a [profession]". Empirically, the generated images were found to exhibit a strong bias towards certain gender and race, and we attempt to improve the diversity of generated images with the proposed equalization loss in this section. We also demonstrate that our approach can also address spurious correlations beyond social biases.

\vspace{-2mm}
\paragraph{A Single Matrix for Comprehensive Debiasing} Unlike the well-defined targets prevalent in zero-shot classification, the nature of generative models requires a more universal solution. Specifically, we seek to derive a debiasing matrix capable of accommodating any prompt. This matrix could subsequently be treated as a standardized preprocessing step, applied prior to the introduction of the embedding into the generator.

To achieve this, we optimize the equalization loss with positive pairs consisting of an enumeration of ``a photo of a [attribute] [profession]'' where the [attribute] is a member of the set of gender or races and the [profession] is a job title sampled from a training set. For instance, to mitigate gender bias, we adopt $\gS=$ $\{$(``a photo of a male doctor'', ``a photo of a female doctor''), $\cdots$, (``a photo of a male engineer'', ``a photo of a female engineer'') $\}$. By solving the calibration matrix with professions in the training set, we expect the obtained matrix can also mitigate the biases in unseen professions. Note that we optimize the equalization loss \eqreff{eq_equal} without applying the initial orthogonal projection matrix $P_0$. This is because our goal is to balance rather than completely eliminate biased information in the generated images.

\iffalse
\subsection{Measuring the Generalization of Calibration}
\vspace{-2mm}

\begin{wraptable}{r}{4.5cm}
\vspace{-4mm}
\setlength\tabcolsep{3.2pt}
\begin{tabular}{l | c c  | c c }
\toprule
&  \multicolumn{2}{c}{\small \textbf{Gender}} & \multicolumn{2}{c}{\small \textbf{Race}}  \\
&  \small before & \small after & \small before & \small after  \\
 \midrule
train & 0.56 & 0.14 & 0.70 & 0.23
\\
test & 0.54 & 0.16 & 0.69 & 0.26
\\
\bottomrule
\end{tabular}
\vspace{-1mm}
\caption{Difference between embeddings ($\lambda = 500$).}\label{table_generalization}
\end{wraptable} 

In this section, we examine whether minimizing the calibration loss with training prompts can yield a calibration matrix that also works for unseen (testing) professions. In particular, we measure the average L2 difference between the projected embedding $\sum_{(i,j) \in \gS_\textnormal{test}} \left \| P z_i -  P z_j \right\| / |\gS_\textnormal{test}|$ for testing prompts and show the results in Table \ref{table_generalization}. We can see that the calibration matrix successfully minimizes the difference after projection, even for unseen professions. \vjnote{This section seems to be an experiment analysis and is better moved to the next section? Also easier to understand the experimental setting later.} 

\fi

\vspace{-2mm}
\section{Experiments: Generative Models}
\vspace{-2mm}

To evaluate the effectiveness of our approach in the context of generative models, we conducted experiments using the Stable Diffusion (SD) v2.1 framework \citep{rombach2022high}. We construct a list of professions that consists of 100 job titles with GPT-4 \citep{openai2023gpt} and randomly separate them into 80 training and 20 testing professions. The complete list can be found in appendix \ref{appendix_exp_details}. In alignment with the framework proposed by \citet{karkkainen2019fairface}, we consider the gender attributes of male and female, and racial attributes of White, Asian, Black, Indian, and Latino\footnote{It is essential to recognize gender and race are complex social constructs that cannot be simply reduced to binary or discrete categories. The choice of using binary gender and discrete race attributes in our work was primarily based on the existing literature and benchmark datasets that have commonly adopted this setting for evaluation purposes.}.

\begin{figure*}[t]
\begin{center}   
\includegraphics[width=\linewidth]{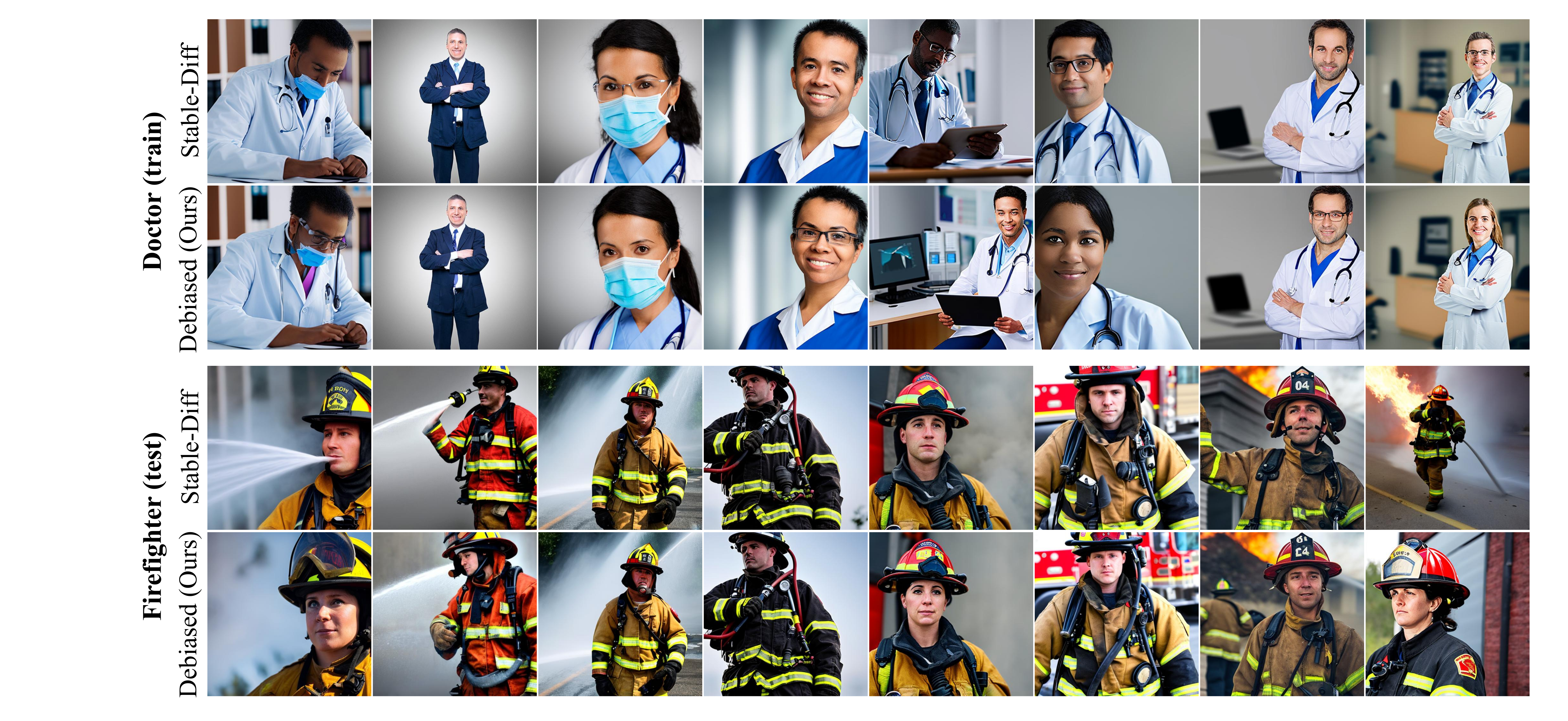}
\end{center}
\caption{\textbf{Improving Gender Diversity of Stable Diffusion.} We fix the random seed of initial latent noise of Stable Diffusion \citep{rombach2022high} and generate the images with the training / testing prompt ``a photo of a doctor / firefighter''. The results demonstrate that applying the calibration matrix to the prompt embedding improves the balance between male and female in the generated images.} \label{fig_gender_sd}
\vspace{-4mm}
\end{figure*}

Evaluating generative models can be challenging without the use of human labels. Inspired by \citet{cho2022dall}, we %employed an empirical approach using 
used sensitive attribute classifiers to predict the sensitive attributes of the generated images. The discrepancy, as defined in \eqref{eq_obj_generative}, was then calculated.  In particular, we generate 100 images for each train / test profession for evaluation, resulting in 10000 images for each model. We then leverage the CLIP classifier to predict the sensitive attributes to calculate the discrepancy. An alternative to CLIP is the FairFace classifier \citep{karkkainen2019fairface}; however, we found that the domain shift between the FairFace dataset and the generated images significantly impairs its performance. The debiased and biased models share the same random seed for fair comparison. We set $\lambda=500$ for all the experiments in this section.

\vspace{-1mm}
\subsection{Measuring the Generalization of Calibration}
\vspace{-1mm}

\begin{wraptable}{r}{4.5cm}
\vspace{-4mm}
\setlength\tabcolsep{3.2pt}
\begin{tabular}{l | c c  | c c }
\toprule
&  \multicolumn{2}{c}{\small \textbf{Gender}} & \multicolumn{2}{c}{\small \textbf{Race}}  \\
&  \small before & \small after & \small before & \small after  \\
 \midrule
train & 0.56 & 0.14 & 0.70 & 0.23
\\
test & 0.54 & 0.16 & 0.69 & 0.26
\\
\bottomrule
\end{tabular}
\vspace{-1mm}
\caption{Difference between embeddings ($\lambda = 500$).}\label{table_generalization}
\end{wraptable} 

We first examine whether minimizing the calibration loss with training prompts can yield a calibration matrix that also works for unseen (testing) professions. In particular, we measure the average L2 difference between the projected embedding $\sum_{(i,j) \in \gS_\textnormal{test}} \left \| P z_i -  P z_j \right\| / |\gS_\textnormal{test}|$ for testing prompts and show the results in Table \ref{table_generalization}. We can see that the calibration matrix successfully minimizes the difference after projection, even for unseen professions.

\subsection{Quantitative and Qualitative Results}

\begin{wraptable}{r}{6.25cm}
\vspace{-4mm}
\setlength\tabcolsep{3.2pt}
\begin{tabular}{r | l | c | c }
\toprule
 &  & \small \textbf{Gender} & \small \textbf{Race}  \\
 \midrule
\mr{2}{\rotatebox[origin=c]{0}{\small Train}} & SD & 0.472$\pm$0.225  & 0.485$\pm$0.160 
\\
 & Ours & \textbf{0.395$\pm$0.205} &   \textbf{0.434$\pm$0.163} 
\\
\midrule
 \mr{2}{\rotatebox[origin=c]{0}{\small Test}} & SD & 0.412$\pm$0.255 & 0.528$\pm$0.184
\\
 & Ours & \textbf{0.354$\pm$0.253} & \textbf{0.455$\pm$0.169}
\\
\bottomrule
\end{tabular}
\caption{\textbf{Discrepancy between Groups}: Calibration matrix reduces the discrepancy over gender and race. The calibration matrix derived from the training set generalizes well to testing set.}\label{table_auto}
\vspace{-2mm}
\end{wraptable} 

The results presented in Table \ref{table_auto} demonstrate a significant reduction in both gender and race discrepancy after debiasing. Importantly, the improvements are observed for both training and testing professions, implying that the obtained debiasing matrix can generalize beyond training prompts. To further illustrate the effectiveness of
our approach, we present quantitative results for mitigating gender bias in Figure \ref{fig_gender_sd}. By applying the calibration matrix to balance the male and female directions, the gender diversity of the generated images significantly improved. Additional examples can be found in appendix \ref{appendix_exp_results}.

\begin{figure*}[t]
\begin{center}   
\includegraphics[width=\linewidth]{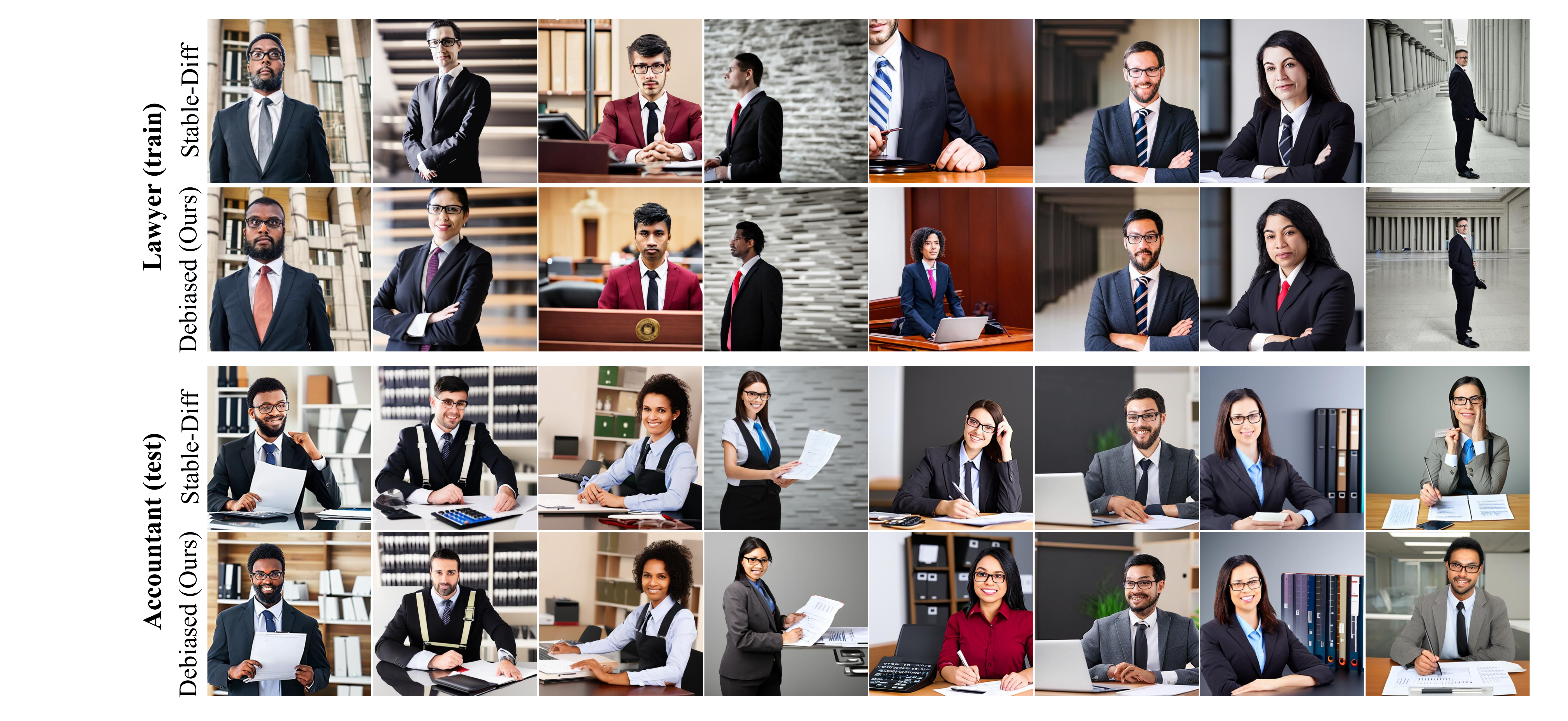}
\end{center}
%\vspace{-3mm}
\caption{\textbf{Improving Racial Diversity of Stable Diffusion.} We again generate the images with Stable Diffusion. After applying the calibration matrix, the race attributes are more diverse in the generated images.} \label{fig_race_sd}
%\vspace{-4mm}
\end{figure*}

Compared to gender bias, we found that addressing racial bias is a more challenging task. One source of complexity is the ambiguity of ethnicity, as individuals may identify with multiple races. Nevertheless, as Figure \ref{fig_race_sd} and Table \ref{table_auto} demonstrate, the diversity in the output images is improved by simply debiasing the prompt embedding with the calibration matrix.

\subsection{Human Evaluation}

\begin{wraptable}{r}{5.6cm}
\vspace{-4mm}
\center
\setlength\tabcolsep{3.4pt}
\begin{tabular}{l | c | c }
\toprule
  & \small \textbf{Gender} & \small \textbf{Race}  \\
 \midrule
SD & 0.472$\pm$0.257  & 0.723$\pm$0.185
\\
 Ours & \textbf{0.372$\pm$0.253} &   \textbf{0.589$\pm$0.188}
\\
\bottomrule
\end{tabular}
\caption{\textbf{Human Evaluation.} We calculate the discrepancy on testing professions with human annotations. Our approach improves the diversity of Stable Diffusion by a non-trivial margin.}\label{table_human}
\end{wraptable}

Despite the scalability, the prediction from a trained classifier could be erroneous. Therefore, we also evaluate our approach with human evaluation, where we invite annotators of different genders, races, and nationalities to label the sensitive attributes of the generated images. Details and the interface are included in appendix \ref{sec_human_eval}. For human evaluation, we generate 25 images for each test profession, resulting in 500 images for each model. As Table
\ref{table_human} shows, our approach greatly improves the diversity of the generated images, corroborating the previous results.

\subsection{Beyond Social Biases}

Our approach can also be applied to address general spurious attributes beyond social biases. As an example, we draw inspiration from the WaterBird dataset \citep{sagawa2019distributionally} and debias the prompt ``a photo of a waterbird'' by using $\{$``a photo of a [animal] with water background'' and ``a photo of a [animal] with land background'' $\}$ as positive pairs, where we construct a list of 100 names of animals with GPT-4 \citep{openai2023gpt}.\begin{wrapfigure}{r}{0.54\textwidth}
  \begin{center}
  \vspace{-4mm}
    \includegraphics[width=0.53\textwidth]{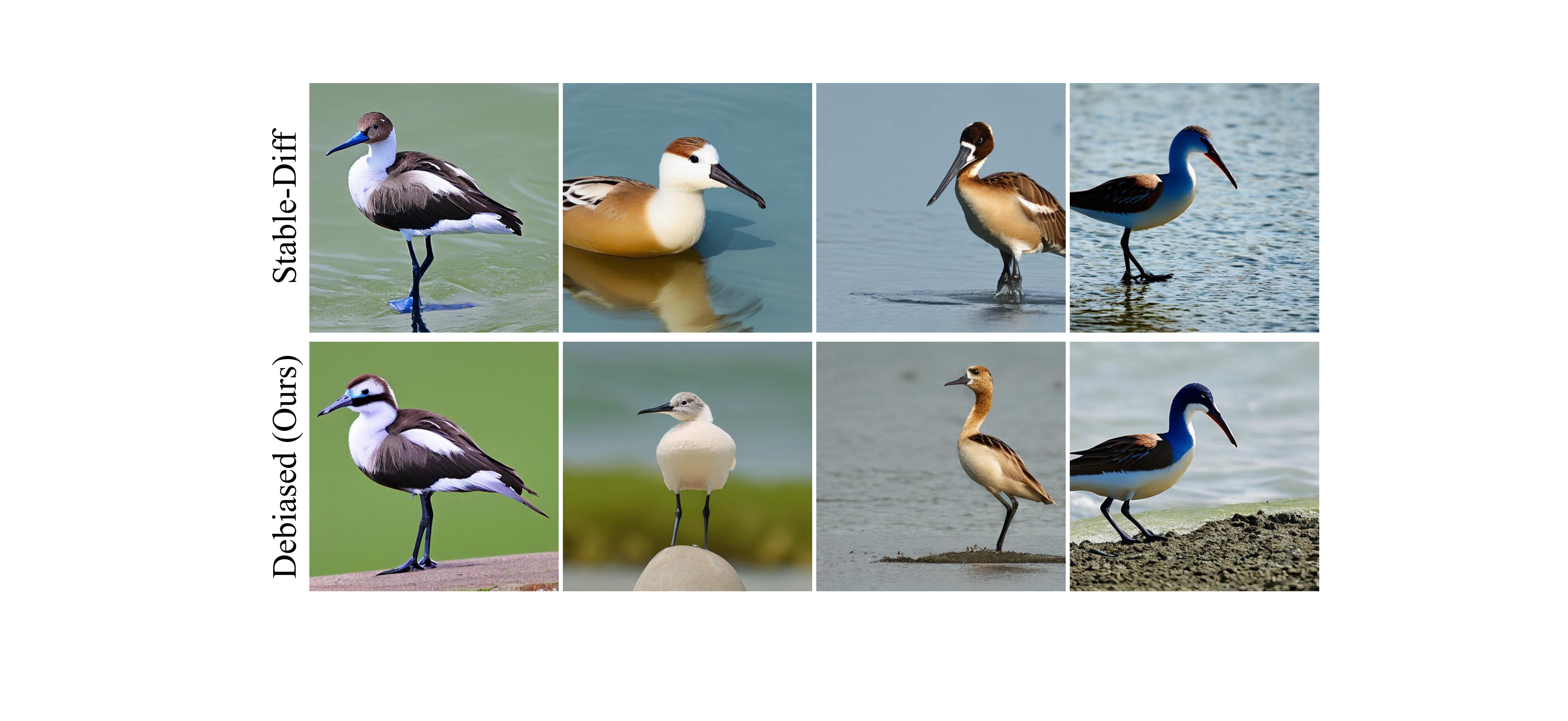}
  \end{center}
  \vspace{-1mm}
  \caption{\textbf{Generation against Non-social Biases.} The results demonstrate the ability of the proposed method to generate images of waterbirds in both land and water backgrounds.} \label{fig_bird}
    \vspace{-10mm}
\end{wrapfigure}
As Figure \ref{fig_bird} illustrates, our approach successfully generates images of water birds in both land and water backgrounds, whereas the original models only generated images with water background.

\vspace{-2mm}
\section{Conclusion}
\vspace{-2mm}

 In this work, we present a new approach to debiasing vision-language foundation models by utilizing prompts to mitigate biases. The proposed calibrated projection effectively mitigates biases in both discriminative and generative vision-language models without any additional training or data. 

\paragraph{Acknowledgements}
Thanks to Arjun Akula, Susanna Rico, Joshua Robinson, Lucy Chai, Kabir Swain, Manel Baradad, Joanna Materzynska, Shobhita Sundaram, Pei-Ling Chiang, and Yi-Yi Chu for their helpful comments and suggestions. This work was in part supported by NSF BIGDATA IIS- 1741341, NSF CAREER 1553284, and NSF AI Institute TILOS. CYC is supported by an IBM PhD Fellowship.

\bibliographystyle{plainnat}
\bibliography{bibfile}

\newpage
\appendix
\section{Broader Impact}

The development and implementation of debiasing techniques in vision and language models has the potential to significantly impact a wide range of industries and applications. By reducing the biases in the models, they will be better able to accurately recognize and understand diverse individuals and groups, leading to more fair and equitable decision-making in fields such as education, employment, and law enforcement. Nevertheless, our approach also has limitations. For instance, the proposed debiased technique for generative models does not work for certain classes or biases. Despite the limitations, our work on debiasing vision and language models is a crucial step towards creating more inclusive and fair technology for all.

\section{Proof}
\subsection{Lemma \ref{lemma_calibrate}}
\begin{proof}
We will leverage the first order optimality criteria to derive the solution.
\begin{align*}
\min_{P} \;  \| P - P_0 \|^2 + \frac{\lambda}{|\gS|} \sum_{(i,j) \in \gS} \| P z_i -  P z_j \|^2
\end{align*}
The loss can be written as
\begin{align*}
\gL(P) &= \frac{\lambda}{|\gS|}( P z_i -  P z_j)^T ( P z_i -  P z_j) + (P - P_0)^T (P - P_0)
    \\
    &=  \frac{\lambda}{|\gS|}({z_i}^T P^T P z_i + {z_j}^T P^T P z_j - {z_i}^T P^T P z_j - {z_j}^T P^T P z_i) + 
    (P^TP + P_0^T P_0 - P_0^T P - P^T P_0)
\end{align*}
Setting the derivate w.r.t. P to zero yields:
\begin{align*}
    \frac{\partial \gL(P)}{\partial P} = \frac{\lambda}{|\gS|} \sum_{(i,j) \in \gS} 2( P z_i {z_i}^T +  P z_j {z_j}^T -  P z_i {z_j}^T -  P z_j {z_i}^T)+ (2P - 2P_0) = 0&
    \\
    P(I + \frac{\lambda}{|\gS|} \sum_{(i,j) \in \gS} (z_i {z_i}^T +  z_j {z_j}^T -  z_i {z_j}^T - z_j {z_i}^T))  = P_0&
    \\
    P = P_0 (I + \frac{\lambda}{|\gS|} \sum_{(i,j) \in \gS} (z_i {z_i}^T +  z_j {z_j}^T -  z_i {z_j}^T - z_j {z_i}^T))^{-1}&
    \\
    = P_0 \Bigl(I + \frac{\lambda}{|\gS|} \sum_{(i,j) \in \gS} (z_i - z_j)(z_i - z_j)^T\Bigl)^{-1} \quad\quad\quad\quad\quad&
\end{align*}

One can also rewrite the objective as
\begin{align*}
&\min_{P} \;  \| P - P_0 \|^2 + \frac{\lambda}{|\gS|} \sum_{(i,j) \in \gS} \| P (z_i -   z_j) \|^2
\\
=& \min_{P} \;  \| P - P_0 \|^2 + \frac{\lambda}{|\gS|} \| P Z_{\textnormal{diff}} \|^2
\end{align*}
We then have 
\begin{align*}
    \frac{\partial \gL(P)}{\partial P} = 2(P - P_0) + 2 \frac{\lambda}{|\gS|} P Z_{\textnormal{diff}} Z_{\textnormal{diff}}^T &= 0
    \\
    P(I + \frac{\lambda}{|\gS|} Z_{\textnormal{diff}} Z_{\textnormal{diff}}^T) &= P_0
    \\
    P &= P_0 (I + \frac{\lambda}{|\gS|} Z_{\textnormal{diff}} Z_{\textnormal{diff}}^T)^{-1}.
\end{align*}
Note that two optimums are equivalent, where the second one is simply the matrix form of the first.
\end{proof}

\iffalse
\subsection{Lemma \ref{lemma_equivalent}}
\begin{proof}
Similarly, the objective can be rewritten as 
\begin{align*}
 &\min_{z} \;  \| z - z_0 \|^2 + \lambda \sum_{(i,j) \in \gS} ( z^T (z_i - z_j) )^2
 \\
 =&\min_{z} \;  \| z - z_0 \|^2 + \lambda \| z^T z_{\textnormal{diff}} \|^2.
\end{align*}
The derivative is:
\begin{align*}
    \frac{\partial \gL(z)}{\partial z} &= 2 z - 2 z_0 + 2 \lambda z_{\textnormal{diff}} z_{\textnormal{diff}}^T z = 0
    \\
    z &= (I + \lambda z_{\textnormal{diff}} z_{\textnormal{diff}}^T)^{-1 }z_0
    \\
    &= \Bigl(I + \lambda \sum_{(i,j) \in \gS} (z_i - z_j)(z_i - z_j)^T\Bigl)^{-1} z_0
\end{align*}
\end{proof}
\fi

\subsection{Lemma \ref{lemma_equivalent}}
\begin{proof}
Similarly, the objective can be rewritten as 
\begin{align*}
 &\min_{z} \;  \| z - z_0 \|^2 + \frac{\lambda}{|\gS|} \sum_{(i,j) \in \gS} ( z^T (z_i - z_j) )^2
 \\
 =&\min_{z} \;  \| z - z_0 \|^2 + \frac{\lambda}{|\gS|} \| z^T Z_{\textnormal{diff}} \|^2.
\end{align*}
The derivative is:
\begin{align*}
    \frac{\partial \gL(z)}{\partial z} &= 2 z - 2 z_0 + 2 \frac{\lambda}{|\gS|}  Z_{\textnormal{diff}} Z_{\textnormal{diff}}^T z  = 0
    \\
    z &= (I + \frac{\lambda}{|\gS|} Z_{\textnormal{diff}} Z_{\textnormal{diff}}^T)^{-1 } z_0
    \\
    &= \Bigl(I + \frac{\lambda}{|\gS|} \sum_{(i,j) \in \gS} (z_i - z_j)(z_i - z_j)^T\Bigl)^{-1}  z_0
\end{align*}
\end{proof}

\begin{table}[h]
\small
\begin{center}{%
\begin{tabularx}{\textwidth}{l | c }
\toprule
 & \textbf{Class Prompt}
  \\
  \midrule
 $y=0$ & This is a picture of a landbird.
\\
 $y=1$ & This is a picture of a waterbird.
\\
  \midrule
& \textbf{Spurious Prompt}
\\
  \midrule
 & This is a land background. \; This is a picture of a forest. \; This is a picture of a moutain. 
 \\
 &This is a picture of a wood. \; This is a water background. \; This is a picture of an ocean. 
 \\
 &This is a picture of a beach. \; This is a picture of a port. \;
 \\
   \midrule
& \textbf{Positive Pairs}
\\
  \midrule
enumerate of & This is a picture of a landbird with land background. 
\\
& This is a picture of a landbird with water background.
\\
& This is a picture of a landbird in the ocean \; This is a picture of a landbird in the water. \; 
\\
&This is a picture of a landbird in the forest.
  \\
  \midrule
enumerate of & This is a picture of a waterbird with land background. 
\\
& This is a picture of a waterbird with water background.\; 
\\
& This is a picture of a waterbird in the ocean \; This is a picture of a waterbird in the water. \; 
\\
&This is a picture of a waterbird in the forest.
  \\
\bottomrule
\end{tabularx}}
\end{center}
\caption{\textbf{Prompts for WaterBird Dataset.} The spurious prompts are sentences that describe the spurious features, that is, the background of the images. In addition to keywords such as land/water background, we further include terms that describe similar concepts, such as forest or ocean. The positive pairs consist of sentences that describe the same type of bird (landbird/waterbird), while phrases that describe the background are appended afterward.
}
\vspace{-0mm}
\label{table_prompt_waterbird}
%\vspace{-3mm}
\end{table}

\begin{table}[h]
\small
\begin{center}{%
\begin{tabularx}{0.95\textwidth}{l | c }
\toprule
 & \textbf{Class Prompt}
  \\
  \midrule
 $y=0$ & A photo of a celebrity with dark hair.
\\
 $y=1$ & A photo of a celebrity with blond hair.
\\
  \midrule
& \textbf{Spurious Prompt}
\\
  \midrule
 & A photo of a male. \; A photo of a male celebrity. \; A photo of a man. 
 \\
 & A photo of a female. \; A photo of a female celebrity. \; A photo of a woman. 
 \\
   \midrule
& \textbf{Positive Pairs}
\\
  \midrule
& (A photo of a male celebrity with dark hair., A photo of a female celebrity with dark hair.)
\\
& (A photo of a male celebrity with blond hair., A photo of a female celebrity with blond hair.)
  \\
\bottomrule
\end{tabularx}}
\end{center}
\caption{\textbf{Prompts for CelebA Dataset.} We found two positive pairs are sufficient to mitigate the biases for CelebA dataset.
}
\vspace{-0mm}
\label{table_prompt_celebA}
%\vspace{-3mm}
\end{table}

\begin{table}[h]
\small
\begin{center}{%
\begin{tabularx}{0.94\textwidth}{ l | c }
\toprule
Prompt: & A photo of a [CONCEPT] person. 
  \\
  \midrule
 CONCEPT: & good, evil, smart, dumb, attractive, unattractive, lawful, criminal, friendly, unfriendly
\\
\bottomrule
\end{tabularx}}
\end{center}
\caption{\textbf{Prompts for text-image retrieval on FairFace Dataset.} We adopt the 10 training concepts from \cite{berg2022prompt} to construct the prompts for FairFace. These concepts are irrelevant to gender, race, or age, which makes them suitable for evaluating the model biases.
}
\vspace{-0mm}
\label{table_prompt_fairface}
%\vspace{-3mm}
\end{table}

\begin{table}[h]
\small
\begin{center}{%
\begin{tabularx}{\textwidth}{ l | c }
\toprule
GENDER: & A photo of a male [profession]. \; A photo of a female [profession].
  \\
  \midrule
 RACE: & A photo of a white [profession]. \; A photo of a black [profession]. \; A photo of an Asian [profession]. 
 \\
 & \; A photo of an Indian [profession]. \; A photo of a Latino [profession].
\\
\bottomrule
\end{tabularx}}
\end{center}
\caption{\textbf{Prompts for debiasing generative models.} We simply use prompts that describe the gender and race attribute as prompts, where we avoid using ambiguous terms such as white and black as the model could wrongly interpret it as the color of the photo.
}
\vspace{-0mm}
\label{table_prompt_generative}
%\vspace{-3mm}
\end{table}

\section{Experiment Details}
\label{appendix_exp_details}

\subsection{Prompts}
\label{sec_prompts}

In this section, we provide the exact prompt we use for all the experiments in the paper in Table \ref{table_prompt_waterbird}, \ref{table_prompt_celebA}, \ref{table_prompt_fairface}, \ref{table_prompt_generative}.

\subsection{Human Evaluation}
\label{sec_human_eval}

\begin{figure}[h]
\begin{center}   
\includegraphics[width=0.9\linewidth]{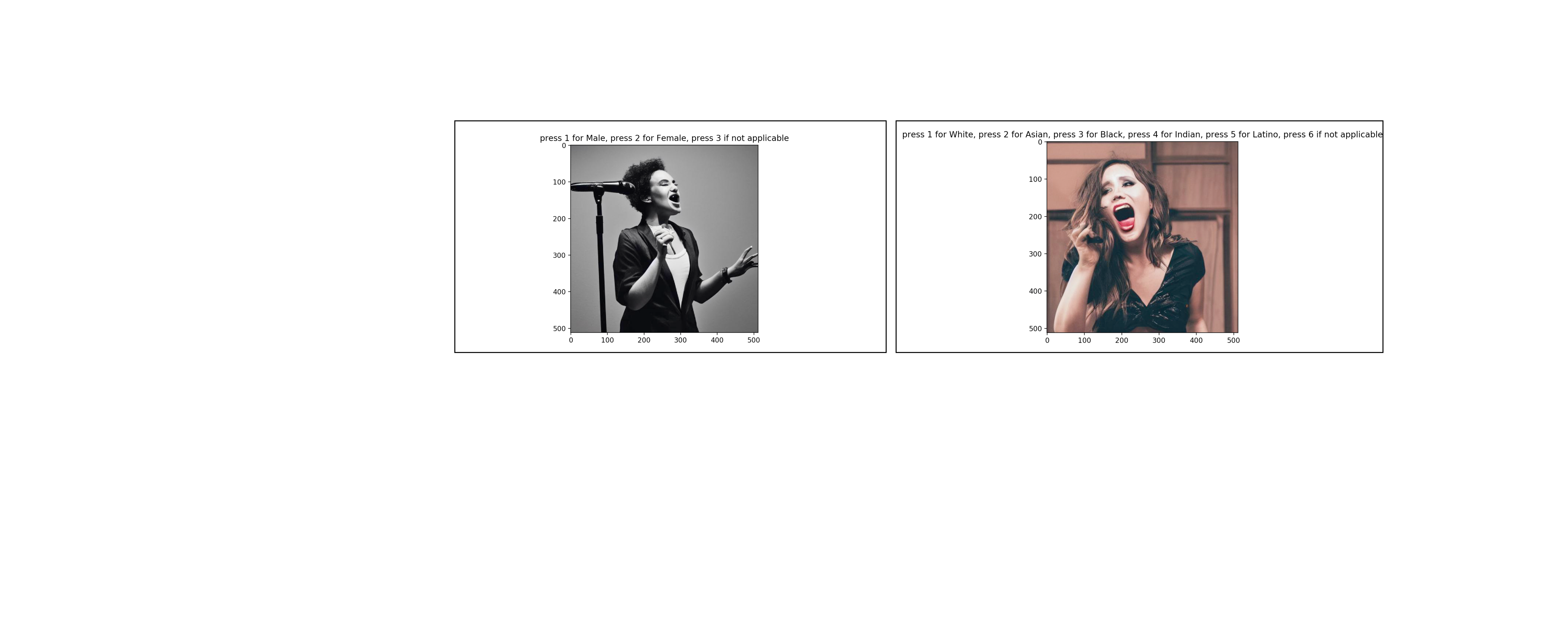}
\end{center}
\vspace{-2mm}
\caption{\textbf{Interface for human evaluation.} We construct a simple labeling script for the annotator to easily label the sensitive attributes. Once they select the label, the program will ask whether they are sure about the answer. One can reselect the answer or press ``enter'' to switch to the next image. } \label{fig_human_eval}
\vspace{-2mm}
\end{figure}

We generate 100 images for each profession for evaluation. Therefore, there are 1500 images for each model in total. The random seed is fixed for the original and debiased Stable Diffusion models. In particular, automatic evaluation and human evaluation adopt the same set of images for fair comparison. The interface for human evaluation is shown in Figure \ref{fig_human_eval}. Note that some generated images might be corrupted, or do not even contain humans. In this case, the annotators can click 3 or 6 to indicate that the current image is not identifiable. We remove these images while calculating the discrepancy.

\section{More Experiment Results}
\label{appendix_exp_results}

\subsection{Importance of Class Name in Positive Pairs}

In this section, we study the importance of class names in the positive pairs. In particular,  instead of using ``a photo of a [class name] with [spurious attribute]'', we instead use ``a photo of a [spurious attribute]'' to estimate the calibration matrix. The results are shown in Table \ref{table_class_imp}. We can see that the performance significantly drops after removing the class name from the prompt, emphasizing the importance of class-conditioned prompts.

\begin{table}[h]
\small
\begin{center}{%
\begin{tabularx}{0.63\textwidth}{l | c c c | c c c }
\toprule
&  \multicolumn{3}{c}{\textbf{Waterbird}} & \multicolumn{3}{c}{\textbf{CelebA}}  \\
 & WG & Avg & Gap & WG & Avg & Gap  \\
 \midrule
Class-Agnostic & 57.5 & 81.4 & 23.9 & 52.8 & 85.2 & 32.4
\\
Class-Conditioned &74.0 & 78.7 & 4.7 & 82.2 & 84.4 & 2.2
\\
\bottomrule
\end{tabularx}}
\end{center}
\caption{\textbf{Importance of target classes in the positive pairs.} Removing [class name] in the positive pairs significantly degenerate the robustness of the zero-shot models.
}
\vspace{-0mm}
\label{table_class_imp}
%\vspace{-3mm}
\end{table}

\subsection{More Samples from Biased and Debiased Generative Models}
\begin{figure}[h]
\begin{center}   
\includegraphics[width=\linewidth]{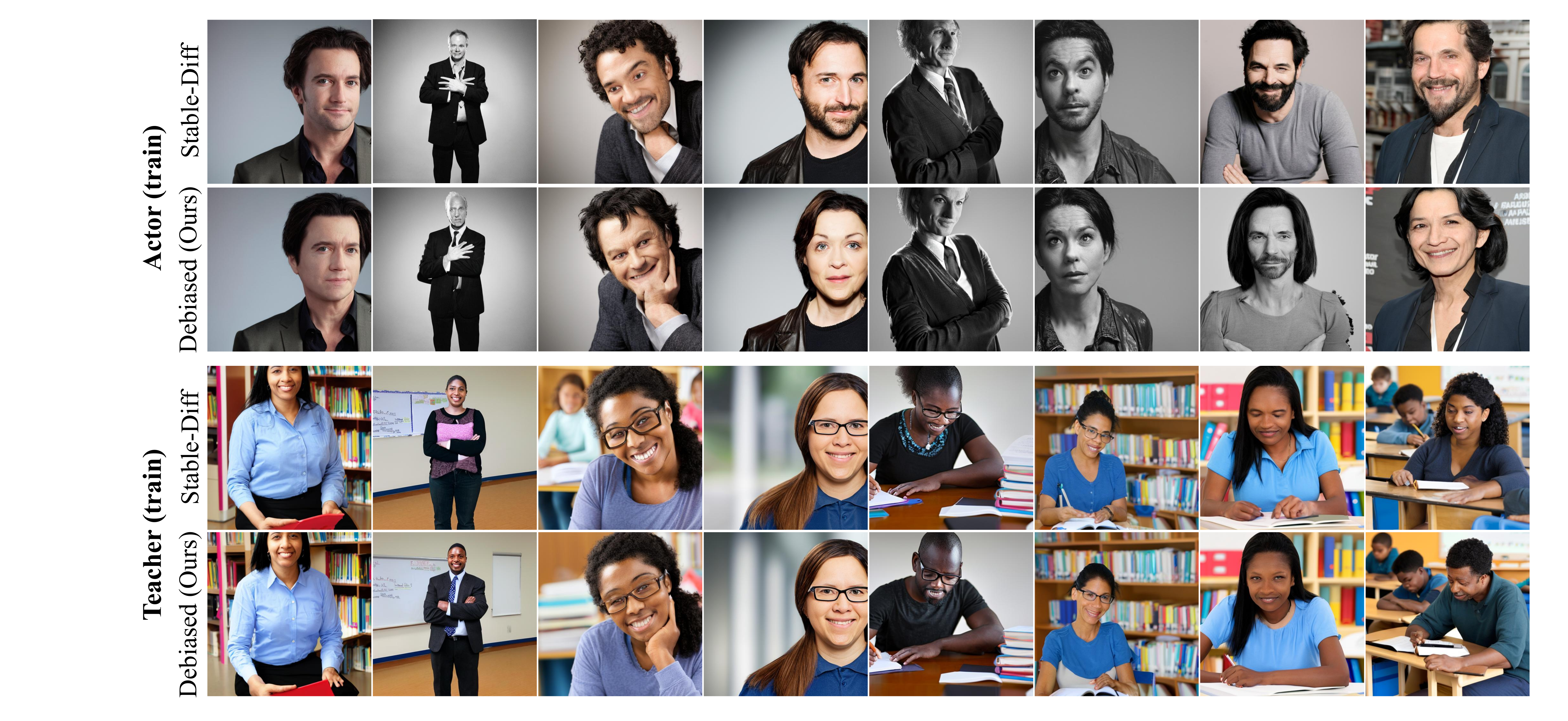}
\end{center}
\vspace{-2mm}
\caption{\textbf{Generation against Gender Bias on Training Set.} After debiasing, we can see that the gender diversity of Stable Diffusion greatly improves. } \label{fig_gender_train}
\vspace{-2mm}
\end{figure}

\begin{figure}[h]
\begin{center}   
\includegraphics[width=\linewidth]{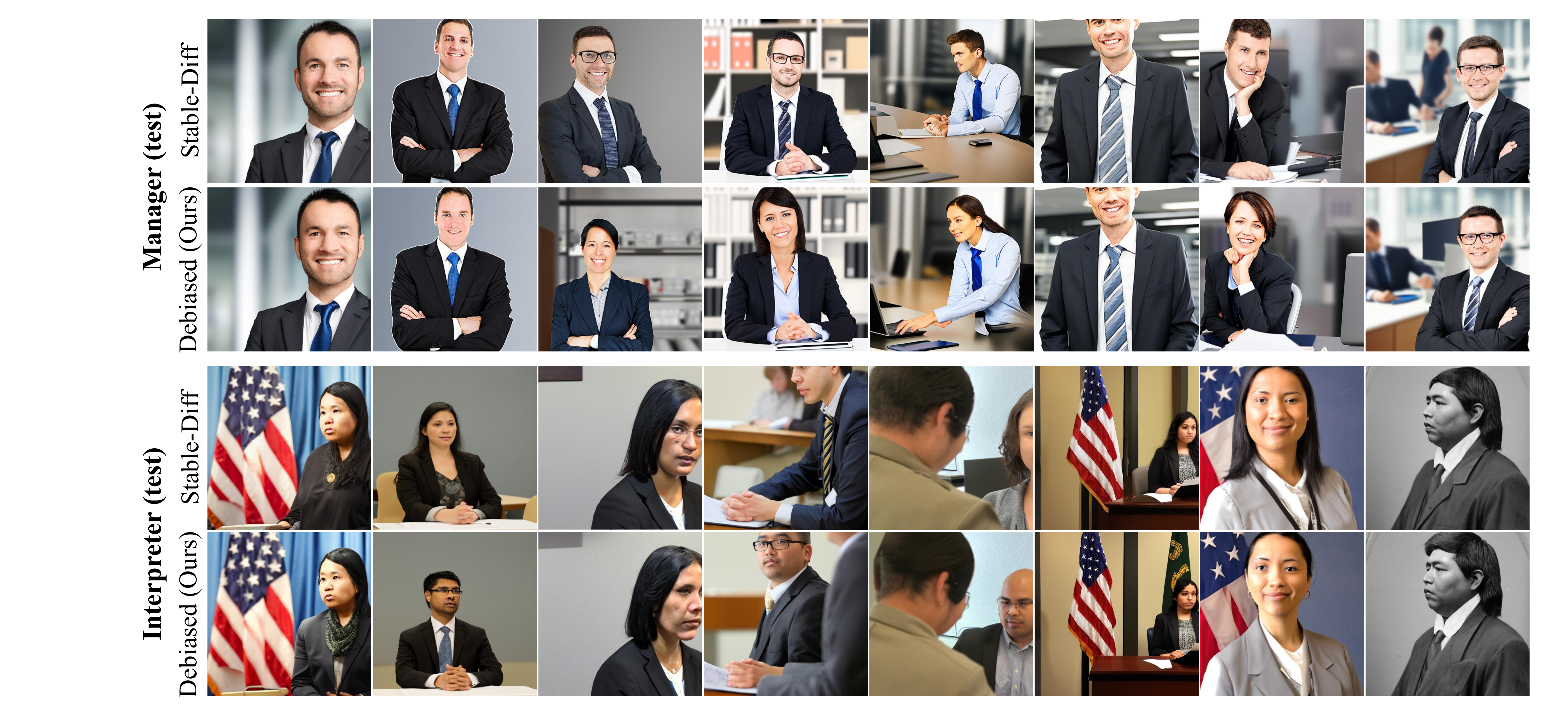}
\end{center}
\caption{\textbf{Generation against Gender Bias on Testing Set.} We can see that by applying the calibration matrix, the gender distributions are more balanced. } \label{fig_gender_test}
\vspace{-4mm}
\end{figure}

\begin{figure}[h]
\begin{center}   
\includegraphics[width=\linewidth]{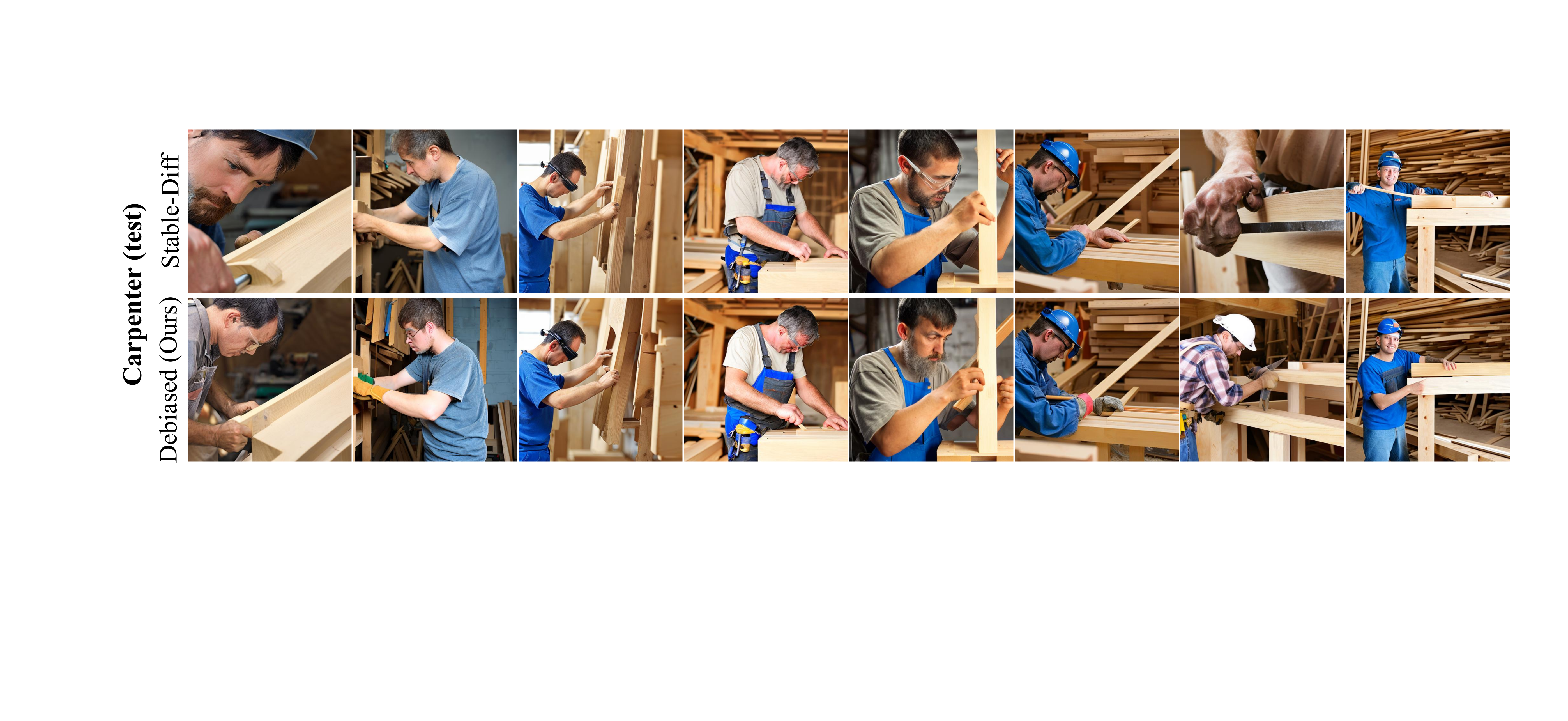}
\end{center}
\caption{\textbf{Failure Case for Generation against Gender Bias.} There are also failure cases, where both our approach and the original model fail. For instance, biased and debiased models fail to generate females for professions such as carpenter and builders. } \label{fig_gender_fail}
\vspace{-4mm}
\end{figure}

\begin{figure}[h]
\begin{center}   
\includegraphics[width=\linewidth]{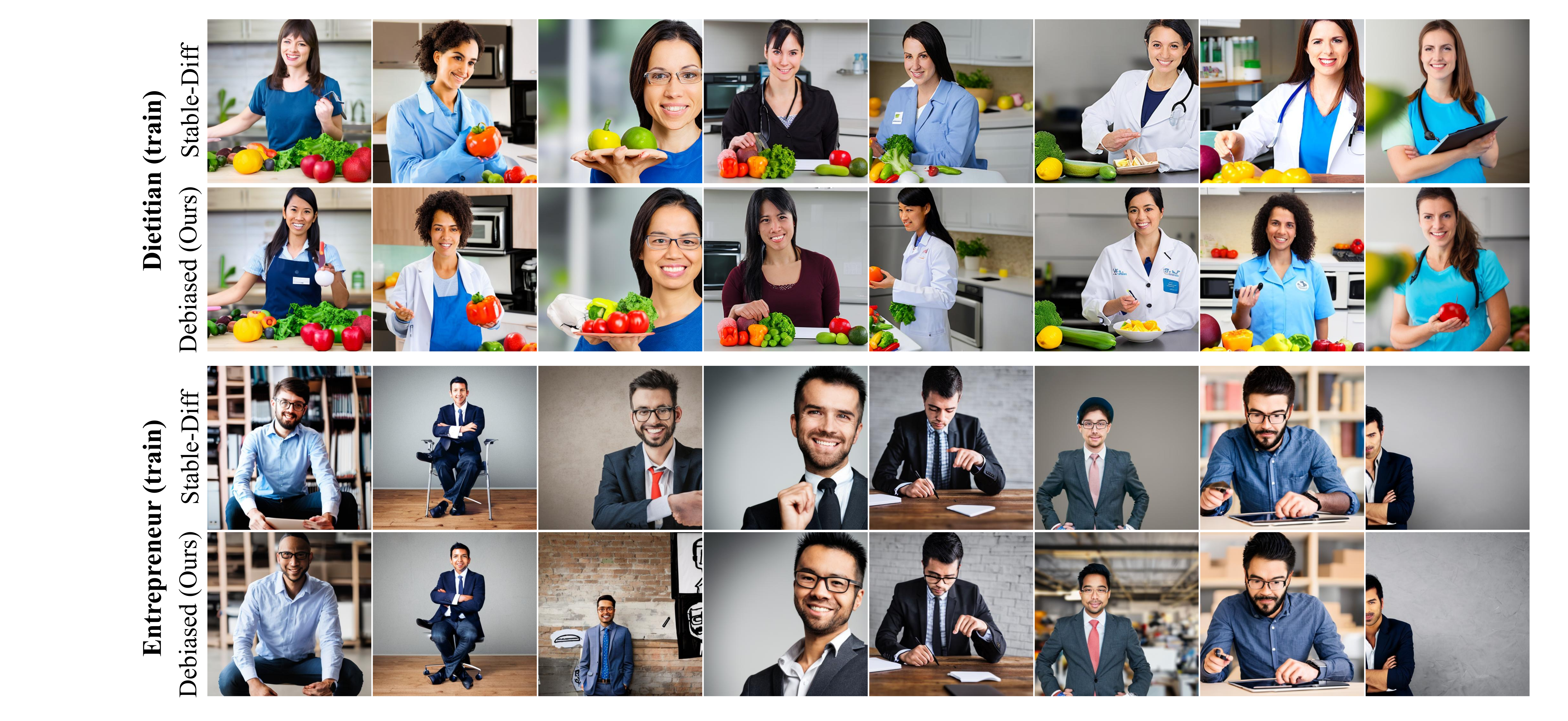}
\end{center}
\caption{\textbf{Generation against Race Bias on Training Set.} We can observe a clear difference before and after debiasing, where the diversity is improved after debiasing.} \label{fig_race_train}
\vspace{-4mm}
\end{figure}

\begin{figure}[h]
\begin{center}   
\includegraphics[width=\linewidth]{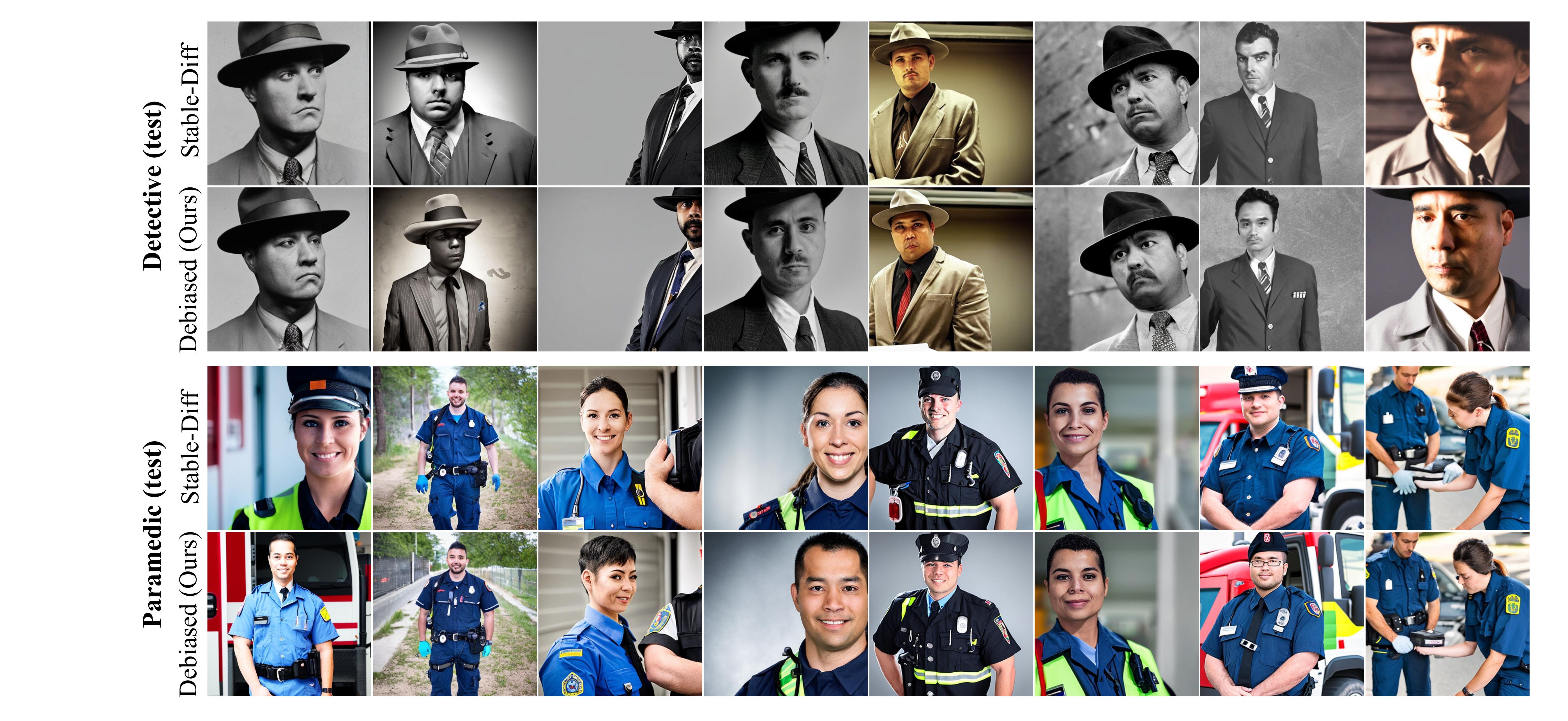}
\end{center}
\caption{\textbf{Generation against Race Bias on Testing Set.} We can again see that the diversity is improved after debising even for unseen classes.} \label{fig_race_test}
\vspace{-4mm}
\end{figure}

In this section, we show more generated images from Stable Diffusion 2.1 to provide a qualitative experiment. We can see that the proposed debiasing approach significantly improves the diverisity across training and testing professions as as Figure \ref{fig_gender_train}, \ref{fig_gender_test}, \ref{fig_race_train} and \ref{fig_race_test} show. Nevertheless, there are also failure cases, where both our approach and the original model fail. For instance, biased and debiased models fail to generate females for many engineer-related professions such as carpenter as Figure \ref{fig_gender_fail} shows.

\begin{figure}[h]
\begin{center}   
\includegraphics[width=\linewidth]{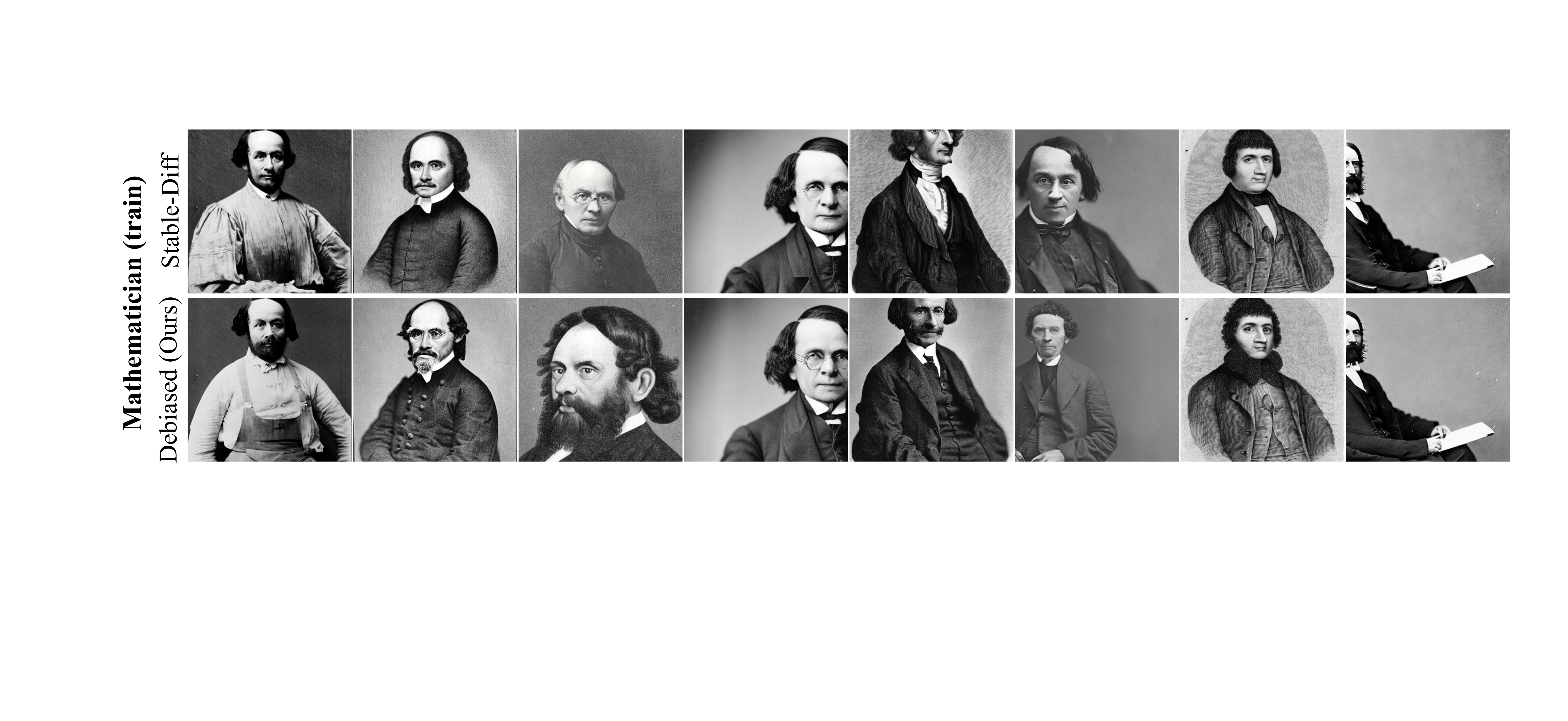}
\end{center}
\caption{\textbf{Failure Case for Generation against Racial Bias.} For certain classes that are highly correlated with historical figures such as mathematicians, our approach does not improve the diversity a lot due to the strong biases from the data. } \label{fig_race_fail}
\vspace{-4mm}
\end{figure}

\begin{table}
\small
\begin{center}{%
\begin{tabularx}{\textwidth}{ l | c }
\toprule
Train & Actor, Architect, Audiologist, Author, Baker, Barber, Blacksmith, Bricklayer
\\
& Bus Driver, Butcher, Chef, Chemist, Cleaner, Coach, Comedian, Computer Programmer
\\
& Construction Worker, Consultant, Counselor, Dancer, Dentist, Designer, Dietitian, DJ
\\
& Doctor, Driver, Economist, Electrician, Engineer, Entrepreneur, Farmer, Florist
\\
& Graphic Designer, Hairdresser, Historian, Journalist, Judge, Lawyer, Librarian, Magician
\\
& Makeup Artist, Mathematician, Marine Biologist, Mechanic, Model, Musician, Nanny, Nurse
\\
& Optician, Painter, Pastry Chef, Pediatrician, Photographer, Plumber, Police Officer, Politician
\\
& Professor, Psychologist, Real Estate Agent, Receptionist, Recruiter, Researcher, Sailor, Salesperson
\\
& Surveyor, Singer, Social Worker, Software Developer, Statistician, Surgeon, Teacher, Technician
\\
& Therapist, Tour Guide, Translator, Vet, Videographer, Waiter, Writer, Zoologist
  \\
  \midrule
 Test & Accountant, Astronaut, Biologist, Carpenter, Civil Engineer, Clerk, Detective
 \\
 & Editor, Firefighter, Interpreter, Manager, Nutritionist, Paramedic, Pharmacist
 \\
 & Physicist, Pilot, Reporter, Security Guard, Scientist, Web Developer
\\
\bottomrule
\end{tabularx}}
\end{center}
\caption{\textbf{100 Training and Testing Professions.} We use GPT-4 to list 100 job titles to form our training and testing set. The similar approach can also be extended to other debiasing tasks by querying large language models.
}
\vspace{-0mm}
\label{table_profession list}
%\vspace{-3mm}
\end{table}

%%%%%%%%%%%%%%%%%%%%%%%%%%%%%%%%%%%%%%%%%%%%%%%%%%%%%%%%%%%%

\end{document}